%% file: acl_latex.tex
\documentclass[11pt]{article}

\usepackage[preprint]{acl}

\usepackage{times}
\usepackage{latexsym}

\usepackage[T1]{fontenc}

\usepackage[utf8]{inputenc}

\usepackage{microtype}

\usepackage{inconsolata}

\usepackage{graphicx}

\usepackage{booktabs}
\usepackage{multirow}
\usepackage{multicol}
\usepackage{amsmath}
\usepackage[inline]{enumitem}
\usepackage{subcaption}
\usepackage[most]{tcolorbox}
\usepackage[OT1]{fontenc}
\usepackage{amssymb}
\usepackage{url}
\usepackage[dvipsnames]{xcolor}
\newcommand{\red}[1]{\textcolor{red}{#1}}
\newcommand{\blue}[1]{\textcolor{blue}{#1}}

\usepackage{tcolorbox}
\usepackage{soul}
\usepackage{hyperref}

\usepackage{amsmath}
\usepackage{amssymb}
\usepackage{pifont}

\definecolor{lightsalmon}{rgb}{1.0, 0.63, 0.48}
\definecolor{lightsalmonpink}{rgb}{1.0, 0.6, 0.6}

\newtcolorbox{hs}[3][]
{
  colframe = black,
  colback  = white,
  coltitle = #2!20!black,  
  sharp corners = southwest,
  arc=2mm,
  boxrule=1pt,
  left=2pt,
  right=2pt,
  top=2pt,
  bottom = 2pt,
  middle = 2pt,
  #1,
}

\newtcolorbox{hsreply}[3][]
{
  before skip=5pt,
  after skip=5pt,
  colframe=black,
  colback=white,
  coltitle=#2!20!black,
  sharp corners=southwest,
  arc=2mm,
  boxrule=1pt,
  left=2pt,
  right=2pt,
  top=2pt,
  bottom=2pt,
  middle=2pt,
  #1,
}

\newtcolorbox{csreply}[3][]
{
  before skip=5pt,
  after skip=5pt,
  colframe=black,
  colback=white,
  coltitle=#2!20!black,
  sharp corners=southeast,
  arc=2mm,
  boxrule=1pt,
  left=2pt,
  right=2pt,
  top=2pt,
  bottom=2pt,
  middle=2pt,
  #1,
}

\title{Deconstructing Stereotypes: Scope-Conditioned Generation for \\ Effective Multilingual Counterspeech}

\author{
  \textbf{Greta Damo\textsuperscript{1}},
  \textbf{Elias Urios Alacreu\textsuperscript{2}},
  \textbf{Elena Cabrio\textsuperscript{1}},
  \textbf{Paolo Rosso\textsuperscript{2,3}},
  \textbf{Serena Villata\textsuperscript{1}}
\\
\\
  \textsuperscript{1}Université Côte d'Azur, CNRS, INRIA, I3S, France \\
  \textsuperscript{2}PRHLT Research Center, Universitat Politècnica de València, Valencia, Spain \\
  \textsuperscript{3}ValgrAI Valencian Graduate School and Research Network of Artificial Intelligence, Spain
\\
  \small{
    \textbf{Correspondence:} \href{mailto:greta.damo@univ-cotedazur.fr}{greta.damo@univ-cotedazur.fr}
  }
}

\begin{document}
\maketitle
\begin{abstract}
Counterspeech (CS) – direct responses that counter online Hate Speech (HS) using reasoning and alternative viewpoints – has emerged as an alternative to content removal. Current automatic CS generation methods, however, frequently produce generic, ineffective replies that fail to target the implicit stereotypes behind HS.
To bridge this gap, we propose a novel scope-conditioned generation framework that explicitly integrates structured stereotype characteristics into Large Language Models prompts. We validate our approach on a novel, human-curated dataset annotated in English, Italian, and Spanish. Extensive evaluations show that stereotype-conditioned prompting substantially outperforms generic baselines across all three languages, obtaining significant gains in factuality, specificity, cogency, and effectiveness for both explicit and implicit implied stereotypes.
\end{abstract}

\noindent \textcolor{red}{Content warning: this paper contains HS examples some readers may find offensive.}

\section{Introduction}

Hate Speech (HS) can be broadly defined as any communication that disparages an individual or group based on protected characteristics such as race, ethnicity, gender, sexual orientation, religion, or nationality \citep{nockleby2000hatespeech}. Driven by the rapid expansion of social media platforms, HS has evolved from a marginal phenomenon into a pervasive one. The Natural Language Processing (NLP) community has responded to this challenge primarily through detection techniques \citep{fortuna2018survey, rawat2024hate}, flagging and removing hateful content at scale. While detection offers an essential first line of defense, it addresses only the surface manifestation of the problem. Content moderation removes the message but leaves the underlying biases untouched, allowing hateful narratives to resurface across platforms or through other users. Consequently, a growing body of work focuses on \textit{Counterspeech} (CS) generation, aiming to directly refute, de-escalate, or reframe hateful content through constructive dialogue \citep{benesch2014countering, bonaldi2024nlp}. 
A foundational yet often under-explored driver of HS is its grounding in \textit{stereotypes}: deeply ingrained cognitive constructs that skew social perception, reinforce inequalities, and justify discriminatory behavior \citep{augoustinos1998construction}. Repeated exposure to stereotypes normalizes bias, escalating over time into overt prejudice and fueling HS \citep{cignarella2025survey}. This connection is core to how HS works, in fact hateful messages persuade by relying on false beliefs the audience already accepts. 


Effective CS must therefore target not only the surface text of a message, but also the underlying stereotype on which it rests. 
However, stereotypes present differently across HS: while some messages state the stereotype explicitly (e.g., claiming \textit{``Women have it so easy; they stay home and have babies''}), others invoke it implicitly through hostility or tropes without naming the underlying premise directly (e.g., wishing harm upon a group via \textit{``I hope all LGBT people d*e of AIDS''}, indirectly implying that all group members carry the disease).
To operationalize stereotypes for computational modelling, prior work in NLP and CS frequently relies on exposing the unstated premise behind a message. 
This underlying stereotype is usually formalized as an \textit{Implied Statement} (IS), which is a simplified, canonical proposition, structured as $\langle\textit{target group}\rangle + \text{relation} + \langle\textit{attributed quality}\rangle$ \citep{sap-etal-2020-social, akazawa-etal-2023-distilling}.
While the IS formulation makes the implicit belief explicit, thereby aiding generation systems in recognizing the true intent of a hateful post, it relies on a simplified abstraction that discards critical structural nuances. Specifically, it ignores: 
\begin{enumerate*}[label=(\roman*)] \item the \textit{scope} of the generalization (e.g., whether a claim targets an essentialist trait of an entire demographic vs.\ a specific subset); and \item the \textit{attributed trait type} quality being assigned (e.g., whether the claim implies moral corruption vs.\ biological inferiority). \end{enumerate*}
By ignoring these dimensions, existing CS systems risk generating generic or misaligned messages that fail to address the HS core intent. 

To bridge this gap, we introduce a novel, multilingual dataset, annotated across English (EN), Spanish (ES), and Italian (IT) that describes stereotypes using a fine-grained annotation schema beyond standard IS, incorporating all these dimensions.
Using this dataset, our work investigates the following Research Questions: \begin{itemize} \item \textbf{RQ1}: \textit{Can stereotype-conditioned generation improve CS quality over generic baselines and IS-only generation?} \item \textbf{RQ2}: \textit{Can stereotype-conditioned generation remain beneficial for implicit stereotypes, given their higher intrinsic difficulty compared to explicit cases?} \item \textbf{RQ3}: \textit{Does stereotype-structured generation transfer cross-lingually?} \end{itemize}
Our main contributions are threefold: \textbf{1)} We present a \textbf{fine-grained stereotype annotation schema} that decomposes HS into its implicit/explicit stereotype, generalization scope, and attributed trait type.
\textbf{2)} We release a high-quality, human-curated \textbf{dataset across three languages} (\textbf{EN}, \textbf{ES}, \textbf{IT}) bridging HS with stereotype modelling, and CS generation\footnote{A data sample and the guidelines are at \url{https://anonymous.4open.science/r/stereotypecs/README.md}}. \textbf{3)} We demonstrate empirically through automatic, LLM-as-a-Judge and human evaluations that \textbf{conditioning LLM-based CS generation on stereotype structure significantly enhances} specificity, factuality, and cogency across all three languages and across both explicit and implicit stereotypes. 

\section{Related Work}

\noindent \textbf{Stereotypes in Language.}
The study of stereotypes in NLP is a relatively new field, and it is closely connected to HS and bias. In fact, HS often conveys harm through implicit stereotypes expressed via generics and pragmatic implicatures rather than explicit insults \cite{fiske1998stereotyping, leslie2014carving, sap-etal-2020-social}. Because these stereotypical beliefs are resistant to change, effective interventions should target the underlying inferences instead of only the surface-level hateful content \cite{lepoutre2019can, perez2021verbal}. Psychology research has shown that stereotype reduction can be achieved through mechanisms such as exposure to counter-stereotypical examples, perspective-taking, and promoting egalitarian norms, though their effectiveness varies across stereotypes and social groups \cite{dasgupta2001malleability, todd2011perspective, wyer2010salient, forscher2019meta, fitzgerald2019interventions}.

\noindent \textbf{Stereotype Detection in NLP.}
Most NLP research on stereotypes focuses on detection, 
as a supervised classification task. Transformer-based models consistently outperform traditional feature-based approaches across multiple languages and benchmarks \cite{sanguinetti2020haspeede, sanchez2021you, pujari2022reinforcement, BOSCO2023, vargas2023socially, cignarella2024queereotypes, schmeisser2025stereohoax, alacreu2025identification}. Related work also examines stereotypes encoded in language models and developed methods for measuring and mitigating representational and stereotypical bias \cite{bolukbasiman, caliskan2017semantics, zmigrod2019counterfactual, sun2019mitigating, laiyk2026stereotype}. While stereotype detection has advanced considerably, comparatively little work has explored methods for mitigating or countering stereotypes.

\noindent \textbf{Stereotype and CS Datasets.}
Several datasets have been developed for stereotype detection and bias evaluation, including CrowS-Pairs \cite{nangia2020crows}, StereoSet \cite{nadeem2021stereoset}, BBQ \cite{parrish-etal-2022-bbq}, HONEST \cite{nozza2021honest}, and multilingual resources such as StereoHOAX \cite{schmeisser2025stereohoax} and QUEEREOTYPES \cite{cignarella2024queereotypes}. Although these datasets cover diverse bias dimensions, 
they remain predominantly English-centric, with relatively limited multilingual and low-resource language coverage. 
Additionally, several datasets have been created to support CS analysis and generation. These include CS collected from social media \cite{mathew2019thou, garland-etal-2020-countering} as well as expert-curated datasets such as CONAN \cite{chung-etal-2019-conan} and its extensions \cite{fanton-etal-2021-human, bonaldi-etal-2022-human, bonaldi-etal-2025-first}.
 While these resources have enabled automatic CS generation, they 
rarely annotate the underlying stereotypes conveyed by it.

\noindent \textbf{CS and Counter-stereotype Generation.}
CS aims to challenge harmful content through strategies such as empathy, fact-checking, humor, denunciation, and warning of consequences \cite{benesch2014countering}. Building on these, NLP research has explored automatic generation using human-written responses \cite{qian2019benchmark}, diversity- and relevance-oriented decoding \cite{zhu2021generate}, controllable generation for response tone \cite{countergedi}, few-shot prompting 
\cite{ashida2022towards}, argumentative components \cite{bonaldi-etal-2024-safer}, and retrieval-augmented generation for factually grounded CS \cite{wilk2025fact, jiang2025rezg, damo2025beating}. However, these methods primarily address explicit HS and focus on response style, with limited attention to countering the implied stereotypes underlying HS.
Countering stereotypes is closely related to CS but focuses on challenging subtler forms of harmful language, where content removal is often inappropriate. Instead, responses aim to educate the speaker, challenge stereotypical beliefs, and signal to bystanders that such statements should not go unchallenged. However, research on directly countering stereotypes is comparatively limited. Psychological studies show that counter-stereotypical examples, humanization, and factual correction can effectively challenge stereotypical beliefs \cite{finnegan2015counter, prati2016humanizing, porter2021global}. In NLP, only few works explicitly model or generate responses to implied stereotypes studying different countering strategies and evaluating their effectiveness \cite{fraser-etal-2021-understanding, fraser-etal-2023-makes, mun-etal-2023-beyond, allaway2023essentialism, nejadgholi2024challenging}, leaving robust counter-stereotype generation unexplored. 
Overall, while substantial progress has been made in stereotype detection and CS generation, comparatively little work has explored LLM-based CS generation that explicitly targets the implied stereotypes underlying hateful messages.
\section{Data}

As our study focuses on EN, ES, and IT, we augment the Multilingual-MTCONAN-KN dataset \cite{bonaldi-etal-2025-first}, which
extends MT-CONAN \cite{bonaldi-etal-2022-human}. It consists of EN HS/CS pairs targeting multiple protected groups, including Women, People of Color (POC), Migrants, Jews, Muslims, and LGBTQ+ individuals, expanded with 
human translations into several languages, including IT and ES. 
We use the training and development splits, 
as they include both the HS and gold CS responses written by human annotators, which we use as reference in our evaluations. Overall, the dataset contains 496 HS/CS pairs for each language, where the ones for ES and IT are human translations from the EN source. 

\begin{table}[t]
\centering
\footnotesize
\begin{tabular}{lccc}
\toprule
\textbf{Dimension} & \textbf{EN} & \textbf{ES} & \textbf{IT} \\
\midrule
Implicit HS     & 0.683 & 0.615 & 0.900 \\
Stereotype                & 0.540 & 0.681 & 0.547 \\
Generalization scope      & 0.919 & 0.813 & 0.774 \\
Trait type                & 0.848 & 0.758 & 0.886 \\ 
\bottomrule
\end{tabular}
\caption{IAA for English, Spanish, and Italian datasets.}
\label{tab:iaa}
\end{table}

\begin{table}[t]
\centering
\footnotesize
\setlength{\tabcolsep}{6pt}
\renewcommand{\arraystretch}{1.05}
\begin{tabular}{@{}lccc@{}}
\toprule
 & \textbf{EN} & \textbf{ES} & \textbf{IT} \\
\midrule
\multicolumn{4}{@{}l}{\textit{Trait type}} \\
\hspace{1em}Existential/Cultural Threat        & 172 & 156 & 130 \\
\hspace{1em}Biological/Physical Discredit      & 117 & 117 & 109 \\
\hspace{1em}Moral/Criminal Threat              & 113 & 143 & 126 \\
\hspace{1em}Intellectual/Cognitive Inferiority &  88 &  94 &  85 \\
\hspace{1em}Economic/Parasitic Drain           &  75 &  75 &  76 \\
\hspace{1em}Cultural Backwardness              &  24 &  26 &  28 \\
\midrule
\multicolumn{4}{@{}l}{\textit{Generalization scope}} \\
\hspace{1em}Majority      & 339 & 328 & 315 \\
\hspace{1em}Essentialist  &  89 &  98 &  89 \\
\hspace{1em}Universal     &  47 &  44 &  48 \\
\midrule
\multicolumn{4}{@{}l}{\textit{Hate speech \& stereotype}} \\
\hspace{1em}Explicit HS         & 434 & 303 & 379 \\
\hspace{1em}Implicit HS         &  41 & 167 &  73 \\
\hspace{1em}Explicit stereotype & 380 & 384 & 369 \\
\hspace{1em}Implicit stereotype &  95 &  86 &  83 \\
\midrule
\hspace{1em}\textbf{Total HS + stereotype} & \textbf{476} & \textbf{470} & \textbf{453} \\
\bottomrule
\end{tabular}
\caption{Summary statistics of the annotations.} 
\label{tab:summary_statistics}
\end{table}

\subsection{Annotation dimensions}
Each HS message is annotated along four dimensions: implicitness of the HS, presence of a stereotype, generalization scope, and attributed trait type. The last two dimensions are annotated only for messages in which a stereotype has been identified. 

\noindent \textbf{Implicitness of HS.} We annotate whether HS is conveyed directly or through inference. HS is labelled \textit{implicit} when derogatory intent is expressed indirectly through devices such as presuppositions, insinuation, coded language, or metaphor; it is labelled \textit{explicit} when conveyed through direct slurs or unambiguous derogatory statements. This dimension is independent of stereotype presence: a message may contain explicit hate while expressing an implicit group-level attribute, and vice versa.


\noindent \textbf{Stereotype presence.} HS is annotated as containing a stereotype when a target group is identifiable, 
an attribute is assigned to it, and the attribute is generalized beyond an individual. Stereotypes are further labeled as \textit{explicit} or \textit{implicit} depending on whether the claim is directly stated or requires inference. This operationalization follows classical definitions of stereotypes as associations between social categories and fixed characteristics \citep{allport1954nature, dovidio2010prejudice} and prior annotation schemes \citep{schmeisser-nieto-etal-2022-criteria}.


\noindent \textbf{Generalization scope.} We annotate the strength of stereotype attribution by distinguishing \textit{universal} (claims holding without exception, e.g., via explicit universal quantifiers), \textit{majority} (statistical or generic claims), 
and \textit{essentialist} (attributes framed as inherent, biological or fixed) scopes. This distinction is grounded in linguistics and philosophy research on generalizations \citep{leslie2008generics, leslie2017original}, categorical and overgeneralized prejudice \citep{allport1954nature, Hamilton1986StereotypesAS}, and psychological essentialism \citep{medin1989psychological, haslam2000essentialist, gelman2003essential}. 

\noindent \textbf{Attributed trait type.} We categorize stereotypes by the domain of characteristics attributed to the target group using six categories: \textit{Intellectual and Cognitive Inferiority, Moral and Criminal Threat, Economic and Parasitic Drain, Biological and Physical Discredit, Existential and Cultural Threat, and Cultural Backwardness}. Annotators assign the most representative category, using multiple labels only when equally prominent. The taxonomy is grounded in the Stereotype Content Model \citep{fiskescm}, which models stereotype content along the dimensions of warmth and competence, and the ABC Model, considering Agency, Beliefs and Communion \citep{koch2016abc}.


\subsection{IS extraction}

The IS is the latent negative stereotype conveyed by a hateful message. 
We add it to the dataset by extracting it, building on
the methodology of \citet{akazawa-etal-2023-distilling}, with three main differences. First, to support our multilingual setting, we use the multilingual Transformer Encoder-Decoder mBART \citep{liu-etal-2020-multilingual-denoising}, 
Second, we expand the data by incorporating the dataset of \citet{bonaldi-etal-2024-safer}, resulting in $\approx$ 37000 samples. Third, two annotators manually validated all the extracted IS and correct inaccurate or incoherent outputs.
Implementation details, hyperparameters, and results are in Appendix~\ref{ap:is}.

\subsection{Annotation process and IAA results}

To enrich the dataset with stereotype-related information, three native speakers\footnote{Aged between 24 and 29, with backgrounds in Data Science, Computer Science, and Communication studies.} per language (EN, ES, IT) have annotated each HS instance. 
To assess annotation reliability, we compute Inter-Annotator Agreement (IAA) using Krippendorff's $\alpha$. 
The annotation proceeded in three stages: we drafted guidelines with definitions and examples; trained the annotators, who then piloted on 30 messages to surface ambiguities, and refined the guidelines based on the pilot IAA before the full annotation


Table~\ref{tab:iaa} reports IAA across languages, with agreement substantial to good overall \citep{krippendorff2018content}. Generalization scope and trait type\footnote{For trait type, we compute Krippendorff's $\alpha$ as a frequency-weighted macro-average of per-label binary alphas, allowing partial credit for overlapping label sets.}
achieve the highest IAA, indicating that, once a stereotype is detected, annotators consistently identify both its generalization strength and its attributed characteristic. By contrast, stereotype presence shows the lowest agreement, reflecting the subjective nature of identifying stereotypes. Implicit HS exhibits the greatest cross-lingual variation, with higher agreement in IT, 
likely because indirect hateful cues are inherently more difficult to identify than explicit ones. Overall, the results support the reliability of the annotation. 

Gold labels are obtained by majority voting. 
When no majority exists for multi-class dimensions, an expert annotator assigns the final one. Table~\ref{tab:summary_statistics} reports summary statistics. Most messages contain a recoverable stereotype (476 in EN, 470 in ES, and 453 in IT, out of 496), confirming that stereotypical generalizations are pervasive in the dataset. Although the HS messages in the original dataset are human translations of the same EN source, small cross-lingual differences remain, consistent with the lower agreement observed for stereotype presence. We also compute IAA on the gold labels across languages, showing consistent  levels w.r.t. the individual languages (Appendix \ref{app:iaa}).
Across languages, both HS and stereotypes are predominantly explicit, except in ES, where implicit HS is more frequent, mirroring its comparatively lower agreement for this dimension. Majority generalizations are the most common scope, followed by essentialist and universal ones. Finally, Existential and Cultural Threat is the most frequent trait type in all languages, followed by Biological and Physical Discredit, and Moral and Criminal Threat, whereas Cultural Backwardness is consistently the least frequent, indicating that stereotype content is largely preserved across translations.

\section{Experimental Setting}

We generate CS using open-source LLMs to enhance the reproducibility of our experiments. To ensure a fair comparison, we select instruction-tuned models from different model families with a similar parameter size (7–9B parameters). The instruction-tuned variants are used to ensure adherence to our prompts, while all models are evaluated using the same generation parameters. 
To investigate whether language specialization influences CS quality, we include both multilingual and monolingual models\footnote{Models specifications and parameters are in Appendix \ref{app:generation}.} 
for the two low resource languages. The multilingual models are \textbf{Ministral} 
\citep{liu2026ministral3}, \textbf{Llama} \cite{touvron2023llama2openfoundation}, and \textbf{EuroLLM} 
\citep{MARTINS202553}, the latter specifically fine-tuned for European languages. The monolingual models are \textbf{Llamantino} \cite{polignano2026advanced} for IT and \textbf{Salamandra} 
\citep{gonzalezagirre2025salamandratechnicalreport} for ES, since both are based on the Llama architecture, they provide a meaningful comparison with the multilingual Llama model while allowing us to assess the impact of language-specific adaptation. The selected models are representative of open-source LLMs commonly adopted in recent CS generation studies. Each model is evaluated in every language under four experimental conditions for a total of 16 generations for both IT and ES (4 models $\times$ 4 conditions) and 12 generations for EN 
(3 models $\times$ 4 conditions). The experimental settings correspond to different levels of information provided as input:  
\begin{enumerate}
    \item \textbf{HS-only (Condition A):} the CS is generated considering only the HS message as input.
    \item \textbf{HS+IS (Condition B):} the CS is generated considering the HS message and the IS.
    \item \textbf{HS+Annotations (Condition C):} the CS is generated considering the HS message and all available annotations.
    \item \textbf{HS+IS+Annotations (Condition D):} the CS is generated considering the HS message, the extracted IS, and all available annotations.
\end{enumerate}

\subsection{Metrics}
Following \cite{saha-etal-2024-crowdcounter, zubiaga-etal-2024-llm}, we conduct automatic, LLM-based, and human evaluations. 

\paragraph{Automatic metrics.} We evaluate generated CS along three dimensions: lexical and semantic overlap, argumentative stance, and diversity. Lexical and semantic overlap are computed against gold reference CS, while stance consistency is evaluated against the original HS.
Concerning the reference CS, for lexical overlap, we report 
METEOR \cite{banerjee-lavie-2005-meteor}, while for semantic similarity
we use BERTScore F1 \cite{zhangbertscore}. We also measure cosine similarity between sentence embeddings of generated CS and reference HS\footnote{sentence-transformers/paraphrase-multilingual-mpnet-base-v2}. To assess the argumentative stance of generated CS w.r.t. the HS, we compute Natural Language Inference (NLI) scores using a multilingual DeBERTa-based entailment model\footnote{mDeBERTa-v3-base-xnli-multilingual-nli-2mil7}, treating HS as premise and CS as hypothesis. We report the average probability mass assigned to the contradiction and entailment classes, where a higher contradiction score indicates the CS is more explicitly opposing or rebutting the hateful claim, while a higher entailment score suggests the CS agrees with or reinforces it, which is undesirable. Finally, output diversity is measured with 
Repetition Rate (RR) \cite{cettolo2014repetition}, capturing the fraction of exactly duplicated generations in the output set.


\begin{table*}[t]
\centering
\scriptsize
\setlength{\tabcolsep}{4pt}
\renewcommand{\arraystretch}{0.8}

\begin{tabular}{ll cccc cccc cccc}
\toprule
& & \multicolumn{4}{c}{\textbf{EN}} & \multicolumn{4}{c}{\textbf{ES}} & \multicolumn{4}{c}{\textbf{IT}} \\
\cmidrule(lr){3-6} \cmidrule(lr){7-10} \cmidrule(lr){11-14}
\textbf{Model} & \textbf{Metric} & \textbf{A} & \textbf{B} & \textbf{C} & \textbf{D} & \textbf{A} & \textbf{B} & \textbf{C} & \textbf{D} & \textbf{A} & \textbf{B} & \textbf{C} & \textbf{D} \\
\midrule

\multirow{8}{*}{\textbf{Llama}} 
& METEOR ($\uparrow$)        & .173 & .164 & .200 & \textbf{.203} & .195 & .192 & \textbf{.220} & .218 & .160 & .157 & .185 & \textbf{.186} \\
& BScore ($\uparrow$)     & \textbf{.859} & .853 & .851 & .849 & \textbf{.687} & .686 & .680 & .673 & \textbf{.678} & .676 & .674 & .671 \\
& HS-Sim. ($\uparrow$)     & .582 & .562 & \textbf{.635} & .623 & .598 & .580 & \textbf{.662} & .626 & .621 & .603 & .631 & \textbf{.642}\\
& RR ($\downarrow$)     & .002 & .004 & \textbf{.000} & \textbf{.000} & .002 & .070 & \textbf{.000} & .023 & .002 & .002 & \textbf{.000} & \textbf{.000}\\
& NLI-Contr. ($\uparrow$)      & .153 & .179 & .387 & \textbf{.494} & .133 & .124 & .422 & \textbf{.545} & .101 & .111 & .244 & \textbf{.349} \\
& NLI-Ent. ($\downarrow$)      & .225 & .207 & .082 & \textbf{.067} & .329 & .327 & .101 & \textbf{.091 }& .450 & .372 & .193 & \textbf{.139} \\

\midrule

\multirow{8}{*}{\textbf{EuroLLM}} 
& METEOR ($\uparrow$)        & .155 & .187 & \textbf{.192} & .188 & .190 & .182 & \textbf{.218} & .209 & .170 & .164 & \textbf{.186} & .173 \\
& BScore ($\uparrow$)     & \textbf{.858} & .856 & .852 & .840 & \textbf{.686} & .663 & .683 & .653 & .675 & \textbf{.686} & .668 & .652 \\
& HS-Sim. ($\uparrow$)     & .542 & .616 & \textbf{.646} & .602 & .600 & .668 & \textbf{.680} & .640 & .636 & .654 & \textbf{.686} & .682 \\
& NLI-Contr. ($\uparrow$)      & .270 & .226 & .434 & \textbf{.446} & .280 & .237 & \textbf{.351} & \textbf{.351} & .206 & .132 & .437 & \textbf{.469} \\
& NLI-Ent. ($\downarrow$)      & .221 & .293 & .182 & \textbf{.126} & .347 & .324 & .317 & \textbf{.245} & .413 & .489 & .210 & \textbf{.115} \\

\midrule

\multirow{8}{*}{\textbf{Ministral}} 
& METEOR ($\uparrow$)        & .177 & .157 & .180 & \textbf{.183} & .178 & .186 & \textbf{.205} & .198 & .171 & .162 & .166 & \textbf{.174 }\\
& BScore ($\uparrow$)     & \textbf{.845} & .844 & .832 & .838 & \textbf{.663} & .653 & .660 & .659 & \textbf{.660} & .658 & .656 & .652 \\
& HS-Sim. ($\uparrow$)     & .606 & .600 & \textbf{.667} & .662 & .575 & .566 & \textbf{.639} & .625 & .603 & .604 & .627 & \textbf{.632} \\
& RR ($\downarrow$)     & \textbf{.000} & .046 & .008 & .017 & .068 & .043 & .028 & \textbf{.019} & \textbf{.000} & .044 & .022 & .002 \\
& NLI-Contr. ($\uparrow$)      & .516 & .236 & .586 & \textbf{.644} & .242 & .214 & \textbf{.500} & .480 & .433 & .154 & .457 & \textbf{.565} \\
& NLI-Ent. ($\downarrow$)      & .116 & .200 & .094 & \textbf{.073} & .227 & .230 & .119 & \textbf{.165} & .169 & .285 & .154 & \textbf{.123} \\

\midrule

\multirow{8}{*}{\shortstack[l]{\textbf{Salamandra (ES) /}\\\textbf{Llamantino (IT)}}} 
& METEOR ($\uparrow$)        & -- & -- & -- & -- & .171 & .170 & .205 & \textbf{.206} & .128 & .118 & \textbf{.156} & .148 \\
& BScore ($\uparrow$)     & -- & -- & -- & -- & \textbf{.696} & .695 & .687 & .688 & \textbf{.680} & .676 & .675 & .658 \\
& HS-Sim. ($\uparrow$)     & -- & -- & -- & -- & .578 & .576 & .603 & \textbf{.611} & .553 & .524 & \textbf{.558} & .546 \\
& RR ($\downarrow$)     & -- & -- & -- & -- & .038 & .096 & \textbf{.000} & .040 & .002 & .013 & .002 & \textbf{.000} \\
& NLI-Contr. ($\uparrow$)      & -- & -- & -- & -- & .175 & .184 & \textbf{.404} & .395 & .070 & .091 & .180 & \textbf{.294} \\
& NLI-En ($\downarrow$)      & -- & -- & -- & -- & .410 & .398 & \textbf{.231} & .259 & .604 & .501 & .344 & \textbf{.247} \\

\bottomrule
\end{tabular}

\caption{Automatic evaluation results across languages, models, and generation conditions (A–D).}
\label{tab:auto_results}
\end{table*}

\paragraph{LLM-as-a-judge and Human evaluation.} We assess the qualitative properties of the generated CS using both an LLM-as-a-judge framework and human evaluation. Both protocols rate each CS on a 5-point Likert scale along six dimensions: \textit{factuality} (the number of specific facts provided); \textit{specificity} (whether the CS directly addresses the topic, target group, and claim of the HS); \textit{effectiveness} (the estimated likelihood of reducing hateful attitudes, challenging underlying beliefs, and persuading bystanders); \textit{correctness} (grammatical accuracy, syntax, and fluency); \textit{safety} (whether the response maintains a respectful, non-harmful tone that attacks ideas, not individuals); and \textit{cogency} (the strength, logical coherence, and number of arguments used to refute the HS). 
For LLM-as-a-Judge, we use three open-weight models of comparable size (Qwen, Ministral, Llama) and average their scores as the final rating, following related work \cite{zubiaga-etal-2024-llm, bonaldi-etal-2025-first} (details are in Appendix \ref{app:lm_judge}). Although two judge models are also evaluated as generation models, we mitigate potential self-evaluation bias by averaging ratings across three independent judges and assessing agreement through IAA with Krippendorff's $\alpha$ among LLM judges, human annotators, and between both (Appendix \ref{app:iaa}).
For human evaluation, three expert native speakers per language independently evaluate a sample along the same six dimensions and additionally indicate their preferred CS among those generated under the conditions. 
\section{Results}

\subsection{Automatic metrics}

Table~\ref{tab:auto_results} reports automatic evaluation results for the four prompting strategies across EN, ES, and IT. Overall, incorporating stereotype information (C and D) consistently improves the characteristics of the generated CS, across languages and LLMs.

Regarding overlap with reference CS, METEOR shows small but consistent improvements from Condition A to D. Conversely, the baseline (A) achieves slightly higher BERTScore values across languages and LLMs, as incorporating stereotype information (C and D) reduces semantic similarity to the reference CS by generating more novel, targeted responses, rather than reproducing them word by word. Semantic similarity between HS and CS remains stable across prompting strategies, with slight increases for Conditions C and D in IT and marginal variance in EN and ES.
Additionally, RR remains very low across all settings, indicating that the additional information does not produce repetitive generations. The largest differences emerge in the NLI analysis. In all languages, Conditions C and D substantially increase the NLI contradiction score while reducing entailment with the HS. This trend is particularly evident for EN, where the contradiction nearly doubles compared to A and B, and remains consistent for ES and IT. As the goal of CS is to challenge rather than reinforce HS, these findings suggest that stereotype-aware prompting encourages more explicit 
opposition to HS.

Comparing prompting conditions, B, which adds IS, performs similarly to the HS-only baseline (A), with no consistent improvements across metrics. In contrast, C and D, which incorporate stereotype information, consistently improve contradiction scores while maintaining similar semantic similarity and low repetition. Combining stereotype information with IS (D) provides only marginal gains over C, suggesting that
stereotype annotations account for most of the observed improvements. 
Differences between conditions are statistically significant (Appendix \ref{app:stat_sig}).

\subsection{LLM-as-a-Judge Evaluation}

Table~\ref{tab:judge_avg} reports the LLM-as-a-Judge evaluation results, averaged across languages and three judge models (Llama, Ministral, and Qwen) 
(disaggregated results are in Appendix \ref{ap:llm_results_complete}). Overall, providing stereotype-related information to the prompt consistently improves CS quality. In fact, Conditions C and D 
outperform strategies relying only on the HS (A) or HS+IS (B).

\begin{table}[h]
\centering
\scriptsize
\setlength{\tabcolsep}{4pt}
\begin{tabular}{llcccc}
\toprule
\textbf{Lang.} & \textbf{Model} & \textbf{A} & \textbf{B} & \textbf{C} & \textbf{D} \\
\midrule
\multirow{3}{*}{\textbf{EN}}
& Llama
& 3.93 {\tiny$\pm$0.45}
& 3.90 {\tiny$\pm$0.46}
& 4.45 {\tiny$\pm$0.29}
& \textbf{4.49} {\tiny$\pm$0.30} \\
& EuroLLM
& 3.35 {\tiny$\pm$0.49}
& 3.75 {\tiny$\pm$0.51}
& 4.00 {\tiny$\pm$0.42}
& \textbf{4.30} {\tiny$\pm$0.35} \\
& Ministral
& 4.56 {\tiny$\pm$0.32}
& 4.38 {\tiny$\pm$0.39}
& 4.85 {\tiny$\pm$0.16}
& \textbf{4.88} {\tiny$\pm$0.14} \\
\midrule
\multirow{4}{*}{\textbf{ES}}
& Llama
& 3.70 {\tiny$\pm$0.40}
& 3.72 {\tiny$\pm$0.39}
& 4.36 {\tiny$\pm$0.29}
& \textbf{4.72} {\tiny$\pm$0.26} \\
& Salamandra
& 3.23 {\tiny$\pm$0.41}
& 3.31 {\tiny$\pm$0.39}
& 3.77 {\tiny$\pm$0.39}
& \textbf{3.86} {\tiny$\pm$0.38} \\
& EuroLLM
& 3.52 {\tiny$\pm$0.42}
& 3.47 {\tiny$\pm$0.54}
& 3.93 {\tiny$\pm$0.35}
& \textbf{4.06} {\tiny$\pm$0.37} \\
& Ministral
& 4.27 {\tiny$\pm$0.37}
& 4.44 {\tiny$\pm$0.34}
& 4.73 {\tiny$\pm$0.19}
& \textbf{4.67} {\tiny$\pm$0.24} \\
\midrule
\multirow{4}{*}{\textbf{IT}}
& Llama
& 3.61 {\tiny$\pm$0.38}
& 3.64 {\tiny$\pm$0.36}
& 4.27 {\tiny$\pm$0.29}
& \textbf{4.33} {\tiny$\pm$0.26} \\
& Llamantino
& 3.36 {\tiny$\pm$0.36}
& 3.12 {\tiny$\pm$0.34}
& \textbf{3.96} {\tiny$\pm$0.39}
& 3.87 {\tiny$\pm$0.42} \\
& EuroLLM
& 3.58 {\tiny$\pm$0.38}
& 3.48 {\tiny$\pm$0.48}
& 4.00 {\tiny$\pm$0.30}
& \textbf{4.15} {\tiny$\pm$0.30} \\
& Ministral
& 4.48 {\tiny$\pm$0.30}
& 4.30 {\tiny$\pm$0.35}
& 4.61 {\tiny$\pm$0.25}
& \textbf{4.69} {\tiny$\pm$0.21} \\
\bottomrule
\end{tabular}
\caption{Average LLM-as-a-Judge scores (over six evaluation dimensions) by language, model, and condition.}
\label{tab:judge_avg}
\end{table}

Across languages and LLMs, C and D particularly improve factuality, specificity, effectiveness, and cogency. For instance, in IT, Llama's overall score increases from 3.61 in Condition A to 4.27 and 4.33 in C and D, respectively, with similar trends observed for EuroLLM and Ministral. These gains 
suggest that explicit stereotype information helps LLMs identify the underlying harmful assumptions and generate more targeted CS. The same trend holds in ES and EN. 
Comparing C and D, adding the IS provides further but smaller improvements, indicating that incorporation of the stereotype characteristics accounts for most of the observed gains, while the IS provides complementary contextual information. The improvement is particularly visible for Llama and EuroLLM, where Condition D achieves the highest overall score in most settings. In contrast, Condition B shows limited and inconsistent improvements over the HS-only baseline, and sometimes it even slightly decreases performance, 
suggesting that identifying only the IS is insufficient to reliably improve CS generation.
Among the evaluated LLMs, Ministral achieves the highest overall scores across languages. Conversely, monolingual LLMs do not outperform multilingual ones, obtaining the lowest scores in IT and ES. Nevertheless, the relative benefit of stereotype-aware prompting is consistent across all LLMs. Finally, safety and correctness remain high across all conditions, showing that additional stereotype information improves CS quality without compromising safety.
\subsection{Human Evaluation}

\begin{table}[ht]
\centering
\resizebox{\linewidth}{!}{%
\begin{tabular}{@{}lcccc@{}}
\toprule
 & \multicolumn{4}{c}{\textbf{Strategy}} \\ \cmidrule(lr){2-5}
\textbf{Metric} & \textbf{A} & \textbf{B} & \textbf{C} & \textbf{D} \\ \midrule
\multicolumn{5}{c}{\textbf{EN}} \\ \midrule
Factuality & 2.81 $\pm$ 0.78 & 2.37 $\pm$ 0.77 & 4.08 $\pm$ 0.70 & \textbf{4.39} $\pm$ 0.53 \\
Specificity & 4.31 $\pm$ 0.58 & 3.96 $\pm$ 0.66 & \textbf{4.87} $\pm$ 0.22 & 4.83 $\pm$ 0.27 \\
Effectiveness & 3.37 $\pm$ 0.49 & 3.09 $\pm$ 0.47 & 4.33 $\pm$ 0.44 & \textbf{4.40} $\pm$ 0.43 \\
Correctness & \textbf{4.99} $\pm$ 0.07 & 4.97 $\pm$ 0.09 & 4.97 $\pm$ 0.13 & 4.96 $\pm$ 0.15 \\
Safety & 4.61 $\pm$ 0.25 & \textbf{4.65} $\pm$ 0.31 & 4.56 $\pm$ 0.36 & 4.63 $\pm$ 0.26 \\
Cogency & 3.24 $\pm$ 0.54 & 2.79 $\pm$ 0.44 & 4.36 $\pm$ 0.48 & \textbf{4.57} $\pm$ 0.35 \\
Overall & 3.89 $\pm$ 0.35 & 3.64 $\pm$ 0.33 & 4.53 $\pm$ 0.29 & \textbf{4.63} $\pm$ 0.22 \\ \midrule
Preference votes & 6 & 2 & 25 & \textbf{42} \\ 

\midrule
\multicolumn{5}{c}{\textbf{ES}} \\ \midrule
Factuality & 2.69 $\pm$ 0.69 & 2.84 $\pm$ 0.69 & \textbf{3.72} $\pm$ 0.79 & 3.64 $\pm$ 0.83 \\
Specificity & 3.03 $\pm$ 0.69 & 3.11 $\pm$ 0.66 & \textbf{4.31} $\pm$ 0.54 & 4.12 $\pm$ 0.76 \\
Effectiveness & 3.08 $\pm$ 0.50 & 3.12 $\pm$ 0.60 & 4.19 $\pm$ 0.59 & \textbf{4.21} $\pm$ 0.74 \\
Correctness & 4.68 $\pm$ 0.18 & 4.67 $\pm$ 0.19 & 4.87 $\pm$ 0.24 & \textbf{4.88} $\pm$ 0.23 \\
Safety & 4.83 $\pm$ 0.24 & 4.69 $\pm$ 0.42 & \textbf{4.88} $\pm$ 0.23 & 4.85 $\pm$ 0.24 \\
Cogency & 3.23 $\pm$ 0.51 & 3.27 $\pm$ 0.56 & \textbf{4.44} $\pm$ 0.51 & 4.31 $\pm$ 0.57 \\
Overall & 3.57 $\pm$ 0.34 & 3.60 $\pm$ 0.37 & \textbf{4.43} $\pm$ 0.36 & 4.34 $\pm$ 0.46 \\ \midrule
Preference votes & 3 & 8 & 31 & \textbf{33} \\ \midrule
\multicolumn{5}{c}{\textbf{IT}} \\ \midrule
Factuality & 2.45 $\pm$ 0.72 & 2.35 $\pm$ 0.87 & 3.47 $\pm$ 1.12 & \textbf{4.29} $\pm$ 0.90 \\
Specificity & 4.05 $\pm$ 0.50 & 3.88 $\pm$ 0.68 & 4.36 $\pm$ 0.59 & \textbf{4.87} $\pm$ 0.22 \\
Effectiveness & 3.45 $\pm$ 0.47 & 3.16 $\pm$ 0.64 & 4.04 $\pm$ 0.70 & \textbf{4.67} $\pm$ 0.37 \\
Correctness & \textbf{4.87} $\pm$ 0.30 & 4.81 $\pm$ 0.36 & 4.84 $\pm$ 0.35 & 4.79 $\pm$ 0.41 \\
Safety & 4.65 $\pm$ 0.46 & 4.17 $\pm$ 0.76 & 4.61 $\pm$ 0.53 & \textbf{4.76} $\pm$ 0.33 \\
Cogency & 3.44 $\pm$ 0.49 & 2.97 $\pm$ 0.54 & 3.93 $\pm$ 0.78 & \textbf{4.68} $\pm$ 0.28 \\
Overall & 3.82 $\pm$ 0.30 & 3.56 $\pm$ 0.48 & 4.21 $\pm$ 0.56 & \textbf{4.68} $\pm$ 0.26 \\ \midrule
Preference votes & 2 & 0 & 15 & \textbf{58} \\ 

\bottomrule
\end{tabular}%
}
\caption{Average human evaluation scores.}
\label{tab:human_eval}
\end{table}

Table~\ref{tab:human_eval} reports human evaluation results. Three native speakers per language\footnote{They have backgrounds in Computer Science, Data Science, and Communication Studies, are aged between 22 and 29, and are balanced between males and females.} evaluate the CS generated by Ministral, the best performing model in the automatic and LLM-as-a-Judge evaluations, on a stratified random sample of 25 HS messages. Each evaluator compares the four prompting strategies, resulting in 900 evaluations (25 HS $\times$ 4 CS $\times$ 3 annotators $\times$ 3 languages). 

The human evaluation confirms the trends observed in the automatic and LLM-as-a-Judge evaluations. Conditions C and D consistently outperform the HS-only (A) and HS+IS (B) strategies, with the largest gains in factuality, specificity, effectiveness, and cogency, while maintaining consistently high correctness and safety. Condition D achieves the best overall performance, obtaining the highest effectiveness and cogency scores in all three languages and the highest factuality scores in IT and EN, while C performs marginally better on factuality and specificity in ES and EN. However, the differences between C and D are small, suggesting that stereotype annotations account for most of the improvement, with the IS providing a complementary benefit. These results indicate that explicitly modeling stereotype characteristics leads to more informative, targeted, and persuasive CS without compromising safety.
Preference judgments show an even clearer trend. Condition D is voted as the most preferred in every language, followed by C, whereas A and B are rarely preferred. Notably, despite similar average scores of C and D in ES, evaluators still favor D in direct comparisons. The same pattern is more pronounced in IT and EN, where D receives substantially more preference votes than C despite only modest differences in the individual evaluation dimensions.

\section{Discussion}

To further examine stereotype-aware prompting, we qualitatively analyze two examples generated by Ministral (HS examples and related CS in EN, ES, IT in are Appendix \ref{ap:ministral_generations}), our best-performing LLM, targeting women. The first HS contains an explicit sexist stereotype, while the second implicitly conveys HS by denying workplace discrimination against women.
Below, we report the HS with the explicit stereotype, and the CS generated with conditions A and D for IT, translated into EN:

\begin{hsreply}{black}{}{}
\footnotesize
\textbf{HS}:
If a woman says she wants an abortion, she’s a cunt.
\end{hsreply}
\begin{csreply}{black}{}{}
\footnotesize
\textbf{CS$_{A}$}: The decision to terminate a pregnancy does not define a person’s dignity or worth; rather, it is a personal and complex decision that involves physical and emotional health as well as individual circumstances. Every woman has the right to decide for herself without being judged, because moralizing about her choice only perpetuates harmful and violent prejudices against those who are already facing sensitive social and medical challenges. True strength lies in respecting differences, not in imposing labels that humiliate and divide.
\end{csreply}
\begin{csreply}{black}{}{}
\footnotesize
\textbf{CS$_D$}: It is by no means true that anyone who freely makes decisions about their reproductive health is automatically a “morally weak” person or a “criminal.” The choice to have an abortion, when exercised lawfully, is a fundamental right recognized by the law and by international human rights organizations, such as the UN, precisely because it addresses complex and personal situations that go beyond moralistic judgments. The generalization that all women who make this decision are “sluts” or immoral ignores the medical, social, and economic context that often forces these choices: unwanted pregnancies, life-threatening risks,
sexual violence, or conditions of poverty that would otherwise make it impossible to raise a child. It is never a superficial choice, but an act of responsibility and personal dignity. Furthermore, labeling a private decision in this way only fuels prejudice and places blame on those who are already facing enormous challenges. The real issues to be discussed are those related to access to healthcare services, the prevention of unwanted pregnancies, and support for mothers—certainly not who has the courage to defend their own body and life.
\end{csreply}

Both examples show a clear improvement from Conditions A/B to C/D. Although A and B reject hateful content, they remain generic and do not address the underlying stereotype. In contrast, C and D directly challenge the harmful generalization: they refute explicit sexist claims with factual evidence and counter-examples, and address implicit bias by exposing and rebutting the underlying assumption. Rather than simply opposing the HS, they explain why the stereotype is unfounded. 
These examples reflect the quantitative results, where C and D achieve higher factuality, specificity, effectiveness, and cogency scores in both LLM-as-a-Judge and human evaluations. 

They support RQ1 by showing that stereotype conditioning produces more effective CS than generic prompting or IS alone, RQ2 by illustrating its value for implicit stereotypes, and RQ3 by transferring to all three languages. 
Results in Tables \ref{tab:explicit_stereotypes} and \ref{tab:stereotype_implicit} (Appendix \ref{app:lm_judge}) further support RQ2: although messages with implicit stereotypes generally receives lower scores than explicit cases, being inherently more difficult, stereotype-conditioned generation consistently improves performance, showing that stereotype annotations provide effective grounding for addressing implicit harmful assumptions.

Finally, the results support RQ3, as stereotype-conditioned generation consistently improves CS quality across EN, ES, and IT, for both general-purpose and language-adapted LLMs. The gains, particularly in factuality, specificity, and cogency, indicate that stereotype structure provides transferable guidance for multilingual CS generation.

\section{Conclusion}

We introduce a stereotype-aware approach for CS generation that explicitly models the harmful stereotypes underlying HS, and we release a novel data source focused on stereotype characteristics annotations in EN, ES, and IT. Unlike approaches that rely only on the surface form of HS or on its inferred IS, our method conditions generation on structured stereotype information.
We evaluate this approach, using multiple LLMs and complementary automatic, LLM-as-a-Judge, and human evaluations.
Our results show that incorporating stereotype structure consistently improves CS quality over generic and IS-only generation (\textbf{RQ1}). 
Moreover, our analysis shows that these benefits extend to implicit cases, where the underlying stereotype must be inferred (\textbf{RQ2}), and transfer across languages, highlighting the potential of stereotype representations as a language independent source of guidance for CS generation (\textbf{RQ3}).
Overall, we demonstrate that explicitly surfacing the structure of stereotypes provides a promising direction for generating more targeted, informative, and persuasive CS in multilingual settings.

\section*{Limitations}

Our study has some limitations. First, LLMs may exhibit model-specific biases in generation, and our experiments cover only a subset of currently available models. Therefore, although our results provide evidence that the proposed approach is effective for the models evaluated, they should not be interpreted as evidence that the approach generalizes to all LLMs. Evaluating a broader range of models, including models with different architectures, sizes, and training data, would provide a more comprehensive assessment of its robustness.

Second, although LLM-as-a-Judge provides a useful and scalable means of evaluating generated CS, the underlying scoring process is not fully transparent. In particular, when multiple evaluation dimensions are assessed simultaneously, it is unclear whether the assigned scores are independent across dimensions or whether judgments on one dimension may influence those on another. We mitigate this concern by using multiple judge models and reporting agreement among them. 

Third, the effectiveness of CS is inherently subjective and may vary across individuals, cultures, and communities. Consequently, no single evaluation metric can fully capture whether a generated response constitutes effective CS. We address this limitation by combining complementary evaluation sources, including automatic metrics, LLM-based judgments, and human evaluation. Nevertheless, our human evaluation is limited in the number of participants and examples considered. A larger-scale evaluation involving more participants and a broader set of examples would provide stronger evidence, although such an evaluation is considerably more costly. Nonetheless, we ensure reliability across individuals, cultures, and communities by employing native, expert people with diverse backgrounds in terms of education, age, and gender.

Finally, our analysis relies on annotations aggregated through majority voting to construct a gold standard. This approach provides a practical way to obtain a single reference label, but it does not explicitly account for disagreement and subjectivity among annotators, though the IAA is good overall. In addition, we observe differences across languages, suggesting that the effectiveness of the proposed approach may depend on language-specific characteristics. Understanding the sources of these cross-lingual differences and determining whether they arise from the models, the data, or cultural and linguistic factors remains an important direction for future work.

\section{Ethical considerations}
In this study, we propose a method to improve the automatic generation of CS specifically targeting the implied stereotypes conveyed in hateful messages. However, LLM-generated content may be subject to biases and may produce harmful or inaccurate responses. Therefore, we emphasize that LLMs should not be used autonomously for this application and require appropriate human oversight and monitoring.

We are also aware of the potential risks associated with the release of the code and data used in our experiments. To mitigate potential misuse, we plan to make these resources available for research purposes only.

Finally, we acknowledge that exposure to hateful content may negatively affect the well-being of annotators and evaluators. We, therefore, recruited only volunteers, who gave informed consent after being told about the potential risks of the task as well as the aims and expected benefits of the study.



\bibliography{custom}

\appendix

\section{Implied Statements Extraction} \label{ap:is}

We use the \texttt{facebook/mbart-large-50} checkpoint. The choice is motivated by the fact that its monolingual counterpart, BART, was the best performing model in the original work by \citet{akazawa-etal-2023-distilling}. After the extraction two annotators manually validate the IS in the following way: when the HS explicitly identified the target group but the generated IS is incorrect, the IS is revised. When the target group is implicit but recoverable, the HS is minimally rewritten to make the target explicit.
Table \ref{tab:finetuning-hyperparams} reports the hyperparameters of the best fine-tuning setting, and Table \ref{tab:single-multi} reports the results on the test set with the fine-tuned model.

\begin{table}[h]
\centering
\small
\begin{tabular}{lr}
\toprule
\textbf{Hyperparameter} & \textbf{Value} \\
\midrule
Optimizer                     & AdamW \\
Learning rate                 & $5 \times 10^{-5}$ \\
LR schedule                   & Cosine \\
Warmup ratio                  & 0.10 \\
Weight decay                  & 0.01 \\
Max.\ gradient norm           & 1.00 \\
Epochs                        & 5 \\
Batch size         & 32 \\
Max.\ input / target length   & 128 / 64 \\
Random seed                   & 42 \\
\bottomrule
\end{tabular}
\caption{Fine-tuning hyperparameters for mBART.}
\label{tab:finetuning-hyperparams}
\end{table}

\begin{table}[hbtp]
\centering
\small
\begin{tabular}{@{}lcc@{}}
\toprule
Metric  & Single & Multi \\
\midrule
BLEU-2  & .22    & .46   \\
ROUGE-1 & .37    & .58   \\
ROUGE-2 & .20    & .39   \\
ROUGE-L & .37    & .57   \\
\bottomrule
\end{tabular}
\caption{Results on the test after fine-tuning mBART.}
\label{tab:single-multi}
\end{table}

\section{Counterspeech Generation}
\label{app:generation}

We use the following versions of LLMs: \textbf{Ministral} (mistralai/Ministral-3-8B-Instruct-2512), \textbf{EuroLLM} (utter-project/EuroLLM-9B), \textbf{Salamandra} (BSC-LT/salamandra-7b-instruct), \textbf{Llamantino} (swap-uniba/LLaMAntino-3-ANITA-8B-Inst-DPO-ITA), and \textbf{Llama} (meta-llama/Llama-3.1-8B-Instruct), with \texttt{top-p} 0.9, \texttt{repetition\_penalty} 1.1, \texttt{max\_new\_tokens} 300, and \texttt{temperature} 0.01 to enhance reproducibility and adherence to the prompt. The parameters are fixed for all models and conditions. The experiments were done using an A100 GPU. We employ the following prompt where the information provided are added incrementally to the input. In Condition A only point 1. is provided, in B 1. and 2., in C 1., 3., 4., and 5., and D has all the information combined.

\input{prompt_cs}


\section{Statistical significance analysis}
\label{app:stat_sig}

\paragraph{Automatic metrics.} We assess differences between generation conditions using paired Wilcoxon signed-rank tests over individual test instances, with Benjamini--Hochberg correction for multiple comparisons. The main aim is to see whether the improvements from conditions A/B to C/D are significant, therefore we make the following comparisons: A vs C, B vs C, A vs D, and B vs D.
Across the four LLMs and languages, a substantial proportion of the differences between conditions are statistically significant ($p\_{\mathrm{FDR}}<0.05$), although the direction and magnitude of the effects vary across metrics and models.

\begin{table}[t]
\centering
\scriptsize
\setlength{\tabcolsep}{5pt}
\begin{tabular}{@{}llrrrr@{}}
\toprule
\textbf{Model} & \textbf{Metric} &
\textbf{A--D} & \textbf{B--D} &
\textbf{A--C} & \textbf{B--C} \\
\midrule
EN Ministral & Correctness
& 1.000 & $<.001$ & 1.000 & $<.001$ \\
EN Ministral & Safety
& .950 & .950 & n/a & .333 \\
ES EuroLLM & Safety
& .559 & $<.001$ & .001 & $<.001$ \\
ES Ministral & Safety
& .620 & .211 & .780 & .637 \\
IT Ministral & Correctness
& .253 & .019 & .621 & $<.001$ \\
IT Ministral & Safety
& n/a & .101 & .635 & .310 \\
\bottomrule
\end{tabular}
\caption{Holm-corrected $p$-values for the six language$\times$model$\times$metric families containing at least one non-significant contrast. n/a denotes an undefined test because all paired differences are zero.}
\label{tab:app-exceptions}
\end{table}

\paragraph{LLM-as-a-Judge.} We test whether the inclusion of stereotype information significantly affects CS quality. The unit of analysis is the HS message, with the four conditions evaluated on the same messages; thus, all comparisons are paired. For each language, generation model, and metric, we compare the judge-averaged scores using two-sided Wilcoxon signed-rank tests with Pratt's zero method. The four contrasts are A vs C, A vs D, B vs C, and B vs D. $p$-values are Holm--Bonferroni corrected within each language$\times$model$\times$metric family, and effect sizes are reported as matched-pairs rank-biserial correlations $r$. In Table \ref{tab:judge_all}, a condition is marked $^*$ ($^\dagger$) when it significantly underperforms D at corrected $p<.05$. Across the 77 language$\times$model$\times$metric families, 71 have all four contrasts significant, always in the expected direction, with C and D outperforming A and B. No contrast is significant in the opposite direction. Among the 291 significant contrasts, the median rank-biserial correlation is $r=0.89$, with $r\geq0.28$ for the four metrics that are not affected by ceiling effects.
The six families for which at least one contrast is not significant are reported in Table~\ref{tab:app-exceptions}. All concern the two ceiling-prone metrics, correctness and safety, for which scores are already very high ($4.79$--$5.00$ out of $5$). The corresponding mean paired differences are at most $0.01$ in absolute value, indicating that the non-significant results reflect practical indistinguishability rather than evidence against the proposed effect. Two comparisons are undefined because all paired differences are exactly zero. Finally, aggregating over messages, C and D outperform A and B in all 11 language$\times$model cells for the overall score, with an average improvement of $+0.53$ points (range: $+0.26$--$+0.74$). This consistency across languages and models supports the robustness of the observed effect within the evaluated experimental design, without implying generalisation to unseen models.

\section{LLM-as-a-judge}
\label{app:lm_judge}

We employ three LLMs to assign scores to the CS generated under different strategies in all languages. We use Qwen (Qwen/Qwen3.5-9B with thinking disabled), Llama (meta-1199
llama/Llama-3.1-8B-Instruct) and Ministral (mistralai/Ministral-3-8B-Instruct-2512), following similar settings in related work \cite{bonaldi-etal-2025-first, zubiaga-etal-2024-llm}. We set \texttt{temperature=0.01} to ensure deterministic behavior. To compensate the fact that two models are used also in the experiments, we employ three judges and one model which is different and we compare the results by computing IAA between them, which appears good overall, as Table \ref{tab:judge_judge_all} shows. We use the following prompt.

\input{prompt_judge}

\subsection{Complete LLM-as-a-judge results}
\label{ap:llm_results_complete}
Table \ref{tab:judge_all} reports all the disaggregated result scores from the LLM-as-a-Judge evaluation, along with their statistical significance. 

\begin{table*}[htbp]
\centering
\scriptsize
\setlength{\tabcolsep}{4pt}
\begin{tabular}{llccccccc}
\toprule
\textbf{Model} & \textbf{Cond.} & \textbf{Fact.\textsuperscript{$\uparrow$}} & \textbf{Spec.\textsuperscript{$\uparrow$}} & \textbf{Eff.\textsuperscript{$\uparrow$}} & \textbf{Corr.\textsuperscript{$\uparrow$}} & \textbf{Safety\textsuperscript{$\uparrow$}} & \textbf{Cog.\textsuperscript{$\uparrow$}} & \textbf{Overall\textsuperscript{$\uparrow$}} \\
\midrule
\multicolumn{9}{c}{\textbf{EN}}\\
\midrule
\multirow{4}{*}{Llama}
& A & 3.03\,\scriptsize$\pm$0.75$^{*\dagger}$ & 3.67\,\scriptsize$\pm$0.67$^{*\dagger}$ & 3.52\,\scriptsize$\pm$0.53$^{*\dagger}$ & 4.87\,\scriptsize$\pm$0.23$^{*\dagger}$ & 4.98\,\scriptsize$\pm$0.09$^{*\dagger}$ & 3.53\,\scriptsize$\pm$0.66$^{*\dagger}$ & 3.93\,\scriptsize$\pm$0.45$^{*\dagger}$ \\
& B & 2.95\,\scriptsize$\pm$0.77$^{*\dagger}$ & 3.63\,\scriptsize$\pm$0.67$^{*\dagger}$ & 3.49\,\scriptsize$\pm$0.53$^{*\dagger}$ & 4.85\,\scriptsize$\pm$0.24$^{*\dagger}$ & 4.97\,\scriptsize$\pm$0.11$^{*\dagger}$ & 3.50\,\scriptsize$\pm$0.68$^{*\dagger}$ & 3.90\,\scriptsize$\pm$0.46$^{*\dagger}$ \\
& D & 3.79\,\scriptsize$\pm$0.53 & 4.53\,\scriptsize$\pm$0.45 & 4.02\,\scriptsize$\pm$0.38 & 4.99\,\scriptsize$\pm$0.07 & 5.00\,\scriptsize$\pm$0.03 & 4.36\,\scriptsize$\pm$0.47 & 4.45\,\scriptsize$\pm$0.29 \\
& E & \textbf{3.81\,\scriptsize$\pm$0.56} & \textbf{4.57\,\scriptsize$\pm$0.47} & \textbf{4.12\,\scriptsize$\pm$0.40} & \textbf{4.99\,\scriptsize$\pm$0.07} & \textbf{5.00\,\scriptsize$\pm$0.02} & \textbf{4.46\,\scriptsize$\pm$0.50} & \textbf{4.49}\,\scriptsize$\pm$0.30 \\
\midrule
\multirow{4}{*}{EuroLLM}
& A & 2.17\,\scriptsize$\pm$0.68$^{*\dagger}$ & 2.80\,\scriptsize$\pm$0.74$^{*\dagger}$ & 2.91\,\scriptsize$\pm$0.56$^{*\dagger}$ & 4.57\,\scriptsize$\pm$0.33$^{*\dagger}$ & 4.91\,\scriptsize$\pm$0.20$^{*\dagger}$ & 2.72\,\scriptsize$\pm$0.71$^{*\dagger}$ & 3.35\,\scriptsize$\pm$0.49$^{*\dagger}$ \\
& B & 2.67\,\scriptsize$\pm$0.74$^{*\dagger}$ & 3.44\,\scriptsize$\pm$0.74$^{*\dagger}$ & 3.36\,\scriptsize$\pm$0.56$^{*\dagger}$ & 4.77\,\scriptsize$\pm$0.33$^{*\dagger}$ & 4.94\,\scriptsize$\pm$0.28$^{*\dagger}$ & 3.30\,\scriptsize$\pm$0.71$^{*\dagger}$ & 3.75\,\scriptsize$\pm$0.51$^{*\dagger}$ \\
& C & 2.99\,\scriptsize$\pm$0.65 & 3.81\,\scriptsize$\pm$0.67 & 3.60\,\scriptsize$\pm$0.46 & 4.86\,\scriptsize$\pm$0.24 & 4.97\,\scriptsize$\pm$0.14 & 3.73\,\scriptsize$\pm$0.61 & 3.99\,\scriptsize$\pm$0.42 \\
& D & \textbf{3.41\,\scriptsize$\pm$0.60} & \textbf{4.32\,\scriptsize$\pm$0.48} & \textbf{3.89\,\scriptsize$\pm$0.45} & \textbf{4.94\,\scriptsize$\pm$0.18} & \textbf{4.98\,\scriptsize$\pm$0.21} & \textbf{4.22\,\scriptsize$\pm$0.51} & \textbf{4.29}\,\textbf{\scriptsize$\pm$0.35} \\
\midrule
\multirow{4}{*}{Ministral}
& A & 3.79\,\scriptsize$\pm$0.70$^{*\dagger}$ & 4.63\,\scriptsize$\pm$0.41$^{*\dagger}$ & 4.37\,\scriptsize$\pm$0.43$^{*\dagger}$ & 4.99\,\scriptsize$\pm$0.07 & \textbf{5.00\,\scriptsize$\pm$0.00} & 4.58\,\scriptsize$\pm$0.46$^{*\dagger}$ & 4.56\,\scriptsize$\pm$0.32$^{*\dagger}$ \\
& B & 3.45\,\scriptsize$\pm$0.77$^{*\dagger}$ & 4.37\,\scriptsize$\pm$0.55$^{*\dagger}$ & 4.16\,\scriptsize$\pm$0.49$^{*\dagger}$ & 4.98\,\scriptsize$\pm$0.09$^{*\dagger}$ & 5.00\,\scriptsize$\pm$0.05 & 4.30\,\scriptsize$\pm$0.58$^{*\dagger}$ & 4.38\,\scriptsize$\pm$0.39$^{*\dagger}$ \\
& C & 4.46\,\scriptsize$\pm$0.46 & 4.93\,\scriptsize$\pm$0.14 & 4.77\,\scriptsize$\pm$0.31 & 5.00\,\scriptsize$\pm$0.05 & 5.00\,\scriptsize$\pm$0.02 & 4.92\,\scriptsize$\pm$0.16 & 4.85\,\scriptsize$\pm$0.16 \\
& D & \textbf{4.56\,\scriptsize$\pm$0.42} & \textbf{4.95\,\scriptsize$\pm$0.14} & 4.80\,\scriptsize$\pm$0.29 &\textbf{ 5.00\,\scriptsize$\pm$0.03} & \textbf{5.00\,\scriptsize$\pm$0.00} & \textbf{4.94\,\scriptsize$\pm$0.14} & \textbf{4.87}\,\scriptsize$\pm$\textbf{0.14} \\
\midrule
\multicolumn{9}{c}{\textbf{ES}}\\
\midrule
\multirow{4}{*}{Llama}
& A & 2.61\,\scriptsize$\pm$0.63$^{*\dagger}$ & 3.38\,\scriptsize$\pm$0.64$^{*\dagger}$ & 3.27\,\scriptsize$\pm$0.44$^{*\dagger}$ & 4.79\,\scriptsize$\pm$0.26$^{*\dagger}$ & 4.98\,\scriptsize$\pm$0.10$^{*\dagger}$ & 3.18\,\scriptsize$\pm$0.61$^{*\dagger}$ & 3.70\,\scriptsize$\pm$0.40$^{*\dagger}$ \\
& B & 2.64\,\scriptsize$\pm$0.64$^{*\dagger}$ & 3.39\,\scriptsize$\pm$0.64$^{*\dagger}$ & 3.28\,\scriptsize$\pm$0.43$^{*\dagger}$ & 4.82\,\scriptsize$\pm$0.24$^{*\dagger}$ & 4.99\,\scriptsize$\pm$0.06$^{*\dagger}$ & 3.21\,\scriptsize$\pm$0.60$^{*\dagger}$ & 3.72\,\scriptsize$\pm$0.39$^{*\dagger}$ \\
& C & 3.62\,\scriptsize$\pm$0.47 & 4.44\,\scriptsize$\pm$0.49 & 3.91\,\scriptsize$\pm$0.34 & 4.96\,\scriptsize$\pm$0.13 & 5.00\,\scriptsize$\pm$0.05 & 4.23\,\scriptsize$\pm$0.48 & 4.36\,\scriptsize$\pm$0.29 \\
& D & \textbf{3.92\,\scriptsize$\pm$0.47} & \textbf{4.69\,\scriptsize$\pm$0.39} & \textbf{4.17\,\scriptsize$\pm$0.36 }& \textbf{4.98\,\scriptsize$\pm$0.07} & \textbf{5.00\,\scriptsize$\pm$0.02} & \textbf{4.54\,\scriptsize$\pm$0.43} & \textbf{4.55}\,\textbf{\scriptsize$\pm$0.26} \\
\midrule
\multirow{4}{*}{Salamandra}
& A & 2.21\,\scriptsize$\pm$0.54$^{*\dagger}$ & 2.76\,\scriptsize$\pm$0.61$^{*\dagger}$ & 2.78\,\scriptsize$\pm$0.48$^{*\dagger}$ & 4.51\,\scriptsize$\pm$0.33$^{*\dagger}$ & 4.84\,\scriptsize$\pm$0.24$^{*\dagger}$ & 2.58\,\scriptsize$\pm$0.58$^{*\dagger}$ & 3.28\,\scriptsize$\pm$0.41$^{*\dagger}$ \\
& B & 2.22\,\scriptsize$\pm$0.53$^{*\dagger}$ & 2.83\,\scriptsize$\pm$0.60$^{*\dagger}$ & 2.84\,\scriptsize$\pm$0.45$^{*\dagger}$ & 4.49\,\scriptsize$\pm$0.32$^{*\dagger}$ & 4.84\,\scriptsize$\pm$0.26$^{*\dagger}$ & 2.64\,\scriptsize$\pm$0.59$^{*\dagger}$ & 3.31\,\scriptsize$\pm$0.39$^{*\dagger}$ \\
& C & 2.86\,\scriptsize$\pm$0.58 & 3.58\,\scriptsize$\pm$0.59 & 3.40\,\scriptsize$\pm$0.42 & 4.81\,\scriptsize$\pm$0.27 & \textbf{4.96\,\scriptsize$\pm$0.22} & 3.42\,\scriptsize$\pm$0.55 & 3.84\,\scriptsize$\pm$0.39 \\
& D & \textbf{2.89\,\scriptsize$\pm$0.57} & \textbf{3.65\,\scriptsize$\pm$0.57} &\textbf{ 3.44\,\scriptsize$\pm$0.40} & \textbf{4.82\,\scriptsize$\pm$0.29} & 4.95\,\scriptsize$\pm$0.24 & \textbf{3.45\,\scriptsize$\pm$0.53} & \textbf{3.87}\,\scriptsize$\pm$\textbf{0.38} \\
\midrule
\multirow{4}{*}{EuroLLM}
& A & 2.39\,\scriptsize$\pm$0.63$^{*\dagger}$ & 3.14\,\scriptsize$\pm$0.68$^{*\dagger}$ & 3.12\,\scriptsize$\pm$0.47$^{*\dagger}$ & 4.67\,\scriptsize$\pm$0.30$^{*\dagger}$ & 4.96\,\scriptsize$\pm$0.12$^{\dagger}$ & 2.95\,\scriptsize$\pm$0.63$^{*\dagger}$ & 3.54\,\scriptsize$\pm$0.42$^{*\dagger}$ \\
& B & 2.33\,\scriptsize$\pm$0.66$^{*\dagger}$ & 3.21\,\scriptsize$\pm$0.77$^{*\dagger}$ & 2.99\,\scriptsize$\pm$0.59$^{*\dagger}$ & 4.54\,\scriptsize$\pm$0.52$^{*\dagger}$ & 4.79\,\scriptsize$\pm$0.51$^{*\dagger}$ & 3.00\,\scriptsize$\pm$0.68$^{*\dagger}$ & 3.48\,\scriptsize$\pm$0.54$^{*\dagger}$ \\
& C & 2.96\,\scriptsize$\pm$0.50 & 3.75\,\scriptsize$\pm$0.54 & 3.49\,\scriptsize$\pm$0.38 & \textbf{4.83\,\scriptsize$\pm$0.24} & 4.95\,\scriptsize$\pm$0.21 & 3.60\,\scriptsize$\pm$0.54 & 3.93\,\scriptsize$\pm$0.35 \\
& D & \textbf{3.10\,\scriptsize$\pm$0.57} & \textbf{4.02\,\scriptsize$\pm$0.56} & \textbf{3.65\,\scriptsize$\pm$0.43} & \textbf{4.83\,\scriptsize$\pm$0.24} & \textbf{4.97\,\scriptsize$\pm$0.17} & \textbf{3.82\,\scriptsize$\pm$0.57} & \textbf{4.06}\,\scriptsize$\pm$\textbf{0.37} \\
\midrule
\multirow{4}{*}{Ministral}
& A & 3.40\,\scriptsize$\pm$0.71$^{*\dagger}$ & 4.33\,\scriptsize$\pm$0.54$^{*\dagger}$ & 3.99\,\scriptsize$\pm$0.45$^{*\dagger}$ & 4.96\,\scriptsize$\pm$0.13$^{*\dagger}$ & 4.99\,\scriptsize$\pm$0.07 & 4.17\,\scriptsize$\pm$0.58$^{*\dagger}$ & 4.31\,\scriptsize$\pm$0.37$^{*\dagger}$ \\
& B & 3.54\,\scriptsize$\pm$0.68$^{*\dagger}$ & 4.55\,\scriptsize$\pm$0.46$^{*\dagger}$ & 4.19\,\scriptsize$\pm$0.46$^{*\dagger}$ & 4.96\,\scriptsize$\pm$0.11$^{*\dagger}$ & 4.99\,\scriptsize$\pm$0.06 & 4.40\,\scriptsize$\pm$0.50$^{*\dagger}$ & 4.44\,\scriptsize$\pm$0.34$^{*\dagger}$ \\
& C & \textbf{4.21\,\scriptsize$\pm$0.45} & \textbf{4.90\,\scriptsize$\pm$0.20} & \textbf{4.44\,\scriptsize$\pm$0.34} & 4\textbf{.99\,\scriptsize$\pm$0.08} & \textbf{5.00\,\scriptsize$\pm$0.07} & \textbf{4.85\,\scriptsize$\pm$0.24} & \textbf{4.73}\,\scriptsize$\pm$\textbf{0.19} \\
& D & 4.11\,\scriptsize$\pm$0.49 & 4.82\,\scriptsize$\pm$0.30 & 4.40\,\scriptsize$\pm$0.37 & 4.97\,\scriptsize$\pm$0.13 & 4.99\,\scriptsize$\pm$0.07 & 4.75\,\scriptsize$\pm$0.33 & 4.67\,\scriptsize$\pm$0.24 \\
\midrule
\multicolumn{9}{c}{\textbf{IT}}\\
\midrule
\multirow{4}{*}{Llama}
& A & 2.45\,\scriptsize$\pm$0.59$^{*\dagger}$ & 3.30\,\scriptsize$\pm$0.59$^{*\dagger}$ & 3.15\,\scriptsize$\pm$0.43$^{*\dagger}$ & 4.74\,\scriptsize$\pm$0.27$^{*\dagger}$ & 4.96\,\scriptsize$\pm$0.12$^{*\dagger}$ & 3.05\,\scriptsize$\pm$0.58$^{*\dagger}$ & 3.61\,\scriptsize$\pm$0.38$^{*\dagger}$ \\
& B & 2.52\,\scriptsize$\pm$0.60$^{*\dagger}$ & 3.35\,\scriptsize$\pm$0.58$^{*\dagger}$ & 3.19\,\scriptsize$\pm$0.37$^{*\dagger}$ & 4.72\,\scriptsize$\pm$0.28$^{*\dagger}$ & 4.95\,\scriptsize$\pm$0.15$^{*\dagger}$ & 3.10\,\scriptsize$\pm$0.54$^{*\dagger}$ & 3.64\,\scriptsize$\pm$0.36$^{*\dagger}$ \\
& C & 3.51\,\scriptsize$\pm$0.48 & 4.30\,\scriptsize$\pm$0.48 & 3.82\,\scriptsize$\pm$0.33 & 4.96\,\scriptsize$\pm$0.13 & 5.00\,\scriptsize$\pm$0.04 & 4.05\,\scriptsize$\pm$0.48 & 4.27\,\scriptsize$\pm$0.29 \\
& D & \textbf{3.57\,\scriptsize$\pm$0.4}3 & \textbf{4.41\,\scriptsize$\pm$0.45} & \textbf{3.86\,\scriptsize$\pm$0.30} & \textbf{4.96\,\scriptsize$\pm$0.12} & \textbf{5.00\,\scriptsize$\pm$0.02} & \textbf{4.17\,\scriptsize$\pm$0.47} & \textbf{4.33}\,\scriptsize$\pm$\textbf{0.26} \\
\midrule
\multirow{4}{*}{Llamantino}
& A & 2.10\,\scriptsize$\pm$0.57$^{*\dagger}$ & 2.96\,\scriptsize$\pm$0.55$^{*\dagger}$ & 2.90\,\scriptsize$\pm$0.40$^{*\dagger}$ & 4.60\,\scriptsize$\pm$0.31$^{*\dagger}$ & 4.93\,\scriptsize$\pm$0.15$^{*\dagger}$ & 2.69\,\scriptsize$\pm$0.56$^{*\dagger}$ & 3.36\,\scriptsize$\pm$0.36$^{*\dagger}$ \\
& B & 1.89\,\scriptsize$\pm$0.52$^{*\dagger}$ & 2.64\,\scriptsize$\pm$0.48$^{*\dagger}$ & 2.63\,\scriptsize$\pm$0.41$^{*\dagger}$ & 4.36\,\scriptsize$\pm$0.32$^{*\dagger}$ & 4.78\,\scriptsize$\pm$0.26$^{*\dagger}$ & 2.38\,\scriptsize$\pm$0.47$^{*\dagger}$ & 3.12\,\scriptsize$\pm$0.34$^{*\dagger}$ \\
& C & \textbf{3.02\,\scriptsize$\pm$0.70} & \textbf{3.80\,\scriptsize$\pm$0.60} & \textbf{3.49\,\scriptsize$\pm$0.41} & \textbf{4.85\,\scriptsize$\pm$0.23} & \textbf{4.99\,\scriptsize$\pm$0.07} & \textbf{3.56\,\scriptsize$\pm$0.56} & \textbf{3.95}\,\scriptsize$\pm$\textbf{0.39} \\
& D & 2.93\,\scriptsize$\pm$0.75 & 3.69\,\scriptsize$\pm$0.64 & 3.38\,\scriptsize$\pm$0.44 & 4.79\,\scriptsize$\pm$0.26 & 4.97\,\scriptsize$\pm$0.13 & 3.42\,\scriptsize$\pm$0.58 & 3.86\,\scriptsize$\pm$0.42 \\
\midrule
\multirow{4}{*}{EuroLLM}
& A & 2.45\,\scriptsize$\pm$0.58$^{*\dagger}$ & 3.22\,\scriptsize$\pm$0.59$^{*\dagger}$ & 3.14\,\scriptsize$\pm$0.43$^{*\dagger}$ & 4.73\,\scriptsize$\pm$0.28$^{*\dagger}$ & 4.95\,\scriptsize$\pm$0.13$^{*\dagger}$ & 2.99\,\scriptsize$\pm$0.57$^{*\dagger}$ & 3.58\,\scriptsize$\pm$0.38$^{*\dagger}$ \\
& B & 2.38\,\scriptsize$\pm$0.62$^{*\dagger}$ & 3.12\,\scriptsize$\pm$0.72$^{*\dagger}$ & 3.03\,\scriptsize$\pm$0.55$^{*\dagger}$ & 4.60\,\scriptsize$\pm$0.34$^{*\dagger}$ & 4.86\,\scriptsize$\pm$0.26$^{*\dagger}$ & 2.88\,\scriptsize$\pm$0.70$^{*\dagger}$ & 3.48\,\scriptsize$\pm$0.48$^{*\dagger}$ \\
& C & 3.07\,\scriptsize$\pm$0.47 & 3.86\,\scriptsize$\pm$0.47 & 3.55\,\scriptsize$\pm$0.34 & 4.88\,\scriptsize$\pm$0.20 & \textbf{4.99\,\scriptsize$\pm$0.08} & 3.67\,\scriptsize$\pm$0.46 & 4.00\,\scriptsize$\pm$0.30 \\
& D & \textbf{3.16\,\scriptsize$\pm$0.47} & \textbf{4.17\,\scriptsize$\pm$0.46} & \textbf{3.68\,\scriptsize$\pm$0.36} & \textbf{4.91\,\scriptsize$\pm$0.21} & 4.96\,\scriptsize$\pm$0.22 & \textbf{4.00\,\scriptsize$\pm$0.49} & \textbf{4.15}\,\scriptsize$\pm$\textbf{0.30} \\
\midrule
\multirow{4}{*}{Ministral}
& A & 3.58\,\scriptsize$\pm$0.63$^{*\dagger}$ & 4.58\,\scriptsize$\pm$0.41$^{*\dagger}$ & 4.19\,\scriptsize$\pm$0.39$^{*\dagger}$ & \textbf{4.99\,\scriptsize$\pm$0.05} & \textbf{5.00\,\scriptsize$\pm$0.00} & 4.51\,\scriptsize$\pm$0.47$^{*\dagger}$ & 4.47\,\scriptsize$\pm$0.30$^{*\dagger}$ \\
& B & 3.40\,\scriptsize$\pm$0.66$^{*\dagger}$ & 4.33\,\scriptsize$\pm$0.55$^{*\dagger}$ & 3.93\,\scriptsize$\pm$0.38$^{*\dagger}$ & 4.97\,\scriptsize$\pm$0.10$^{*\dagger}$ & 5.00\,\scriptsize$\pm$0.04 & 4.18\,\scriptsize$\pm$0.57$^{*\dagger}$ & 4.30\,\scriptsize$\pm$0.35$^{*\dagger}$ \\
& C & 4.01\,\scriptsize$\pm$0.52 & 4.74\,\scriptsize$\pm$0.32 & 4.29\,\scriptsize$\pm$0.40 & 4.99\,\scriptsize$\pm$0.06 & \textbf{5.00\,\scriptsize$\pm$0.00} & 4.65\,\scriptsize$\pm$0.38 & 4.61\,\scriptsize$\pm$0.25 \\
& D & \textbf{4.15\,\scriptsize$\pm$0.4}9 & \textbf{4.85\,\scriptsize$\pm$0.25} & \textbf{4.35\,\scriptsize$\pm$0.35} & 4.99\,\scriptsize$\pm$0.06 & 5.00\,\scriptsize$\pm$0.05 & \textbf{4.80\,\scriptsize$\pm$0.28} & \textbf{4.69}\,\scriptsize$\pm$\textbf{0.21 }\\
\bottomrule
\end{tabular}
\caption{LLM-as-a-Judge scores across EN, ES, IT. $^*$/$^\dagger$: significantly worse than condition C/D, respectively (two-sided Wilcoxon signed-rank test over items, Holm-corrected within each model$\times$metric family, $p<.05$.}
\label{tab:judge_all}
\end{table*}

\begin{table*}[t]
\centering
\scriptsize
\setlength{\tabcolsep}{4pt}
\begin{tabular}{llccccccc}
\toprule
\textbf{Model} & \textbf{Cond.} & \textbf{Fact.\textsuperscript{$\uparrow$}} & \textbf{Spec.\textsuperscript{$\uparrow$}} & \textbf{Eff.\textsuperscript{$\uparrow$}} & \textbf{Corr.\textsuperscript{$\uparrow$}} & \textbf{Saf.\textsuperscript{$\uparrow$}} & \textbf{Cog.\textsuperscript{$\uparrow$}} & \textbf{Avg.\textsuperscript{$\uparrow$}} \\
\midrule
\multicolumn{9}{c}{\textbf{EN}}\\
\midrule

\multirow{4}{*}{Llama}
& A & 3.10 {\tiny$\pm$ 0.73} & 3.72 {\tiny$\pm$ 0.65} & 3.56 {\tiny$\pm$ 0.51} & 4.88 {\tiny$\pm$ 0.22} & 4.98 {\tiny$\pm$ 0.09} & 3.57 {\tiny$\pm$ 0.65} & 3.97 {\tiny$\pm$ 0.44} \\
& B & 3.02 {\tiny$\pm$ 0.76} & 3.67 {\tiny$\pm$ 0.66} & 3.52 {\tiny$\pm$ 0.53} & 4.86 {\tiny$\pm$ 0.24} & 4.97 {\tiny$\pm$ 0.10} & 3.54 {\tiny$\pm$ 0.68} & 3.93 {\tiny$\pm$ 0.45} \\
& C & 3.83 {\tiny$\pm$ 0.52} & 4.56 {\tiny$\pm$ 0.43} & 4.07 {\tiny$\pm$ 0.37} & \textbf{4.99 {\tiny$\pm$ 0.07}} & \textbf{5.00 {\tiny$\pm$ 0.02}} & 4.41 {\tiny$\pm$ 0.45} & 4.48 {\tiny$\pm$ 0.28} \\
& D & \textbf{3.87 {\tiny$\pm$ 0.55}} & \textbf{4.62 {\tiny$\pm$ 0.44}} & \textbf{4.16 {\tiny$\pm$ 0.39}} & \textbf{4.99 {\tiny$\pm$ 0.07}} & \textbf{5.00 {\tiny$\pm$ 0.02}} & \textbf{4.52 {\tiny$\pm$ 0.47}} & \textbf{4.53} {\tiny$\pm$ \textbf{0.29}} \\
\midrule

\multirow{4}{*}{EuroLLM}
& A & 2.20 {\tiny$\pm$ 0.69} & 2.83 {\tiny$\pm$ 0.75} & 2.93 {\tiny$\pm$ 0.57} & 4.57 {\tiny$\pm$ 0.34} & 4.91 {\tiny$\pm$ 0.20} & 2.75 {\tiny$\pm$ 0.73} & 3.36 {\tiny$\pm$ 0.50} \\
& B & 2.72 {\tiny$\pm$ 0.74} & 3.48 {\tiny$\pm$ 0.75} & 3.40 {\tiny$\pm$ 0.56} & 4.78 {\tiny$\pm$ 0.34} & 4.93 {\tiny$\pm$ 0.30} & 3.35 {\tiny$\pm$ 0.72} & 3.78 {\tiny$\pm$ 0.52} \\
& C & 3.03 {\tiny$\pm$ 0.64} & 3.84 {\tiny$\pm$ 0.66} & 3.62 {\tiny$\pm$ 0.45} & 4.87 {\tiny$\pm$ 0.23} & \textbf{4.98 {\tiny$\pm$ 0.10}} & 3.75 {\tiny$\pm$ 0.60} & 4.02 {\tiny$\pm$ 0.41} \\
& D & \textbf{3.43 {\tiny$\pm$ 0.59}} & \textbf{4.34 {\tiny$\pm$ 0.48}} & \textbf{3.90 {\tiny$\pm$ 0.46}} & \textbf{4.93 {\tiny$\pm$ 0.19}} & 4.98 {\tiny$\pm$ 0.22} & \textbf{4.24 {\tiny$\pm$ 0.51}} & \textbf{4.30} {\tiny$\pm$ \textbf{0.35}} \\
\midrule

\multirow{4}{*}{Ministral}
& A & 3.88 {\tiny$\pm$ 0.67} & 4.69 {\tiny$\pm$ 0.38} & 4.43 {\tiny$\pm$ 0.42} & 4.99 {\tiny$\pm$ 0.07} & \textbf{5.00 {\tiny$\pm$ 0.00}} & 4.65 {\tiny$\pm$ 0.44} & 4.61 {\tiny$\pm$ 0.30} \\
& B & 3.56 {\tiny$\pm$ 0.74} & 4.43 {\tiny$\pm$ 0.52} & 4.21 {\tiny$\pm$ 0.49} & 4.99 {\tiny$\pm$ 0.07} & 5.00 {\tiny$\pm$ 0.02} & 4.37 {\tiny$\pm$ 0.56} & 4.43 {\tiny$\pm$ 0.37} \\
& C & 4.51 {\tiny$\pm$ 0.44} & 4.94 {\tiny$\pm$ 0.14} & 4.79 {\tiny$\pm$ 0.30} & 4.99 {\tiny$\pm$ 0.06} & 5.00 {\tiny$\pm$ 0.02} & 4.93 {\tiny$\pm$ 0.16} & 4.86 {\tiny$\pm$ 0.15} \\
& D & \textbf{4.60 {\tiny$\pm$ 0.41}} & \textbf{4.96 {\tiny$\pm$ 0.13}} & \textbf{4.81 {\tiny$\pm$ 0.27}} & \textbf{5.00 {\tiny$\pm$ 0.02}} & \textbf{5.00 {\tiny$\pm$ 0.00}} & \textbf{4.95 {\tiny$\pm$ 0.13}} & \textbf{4.89} {\tiny$\pm$ \textbf{0.13}} \\

\midrule
\multicolumn{9}{c}{\textbf{ES}}\\
\midrule

\multirow{4}{*}{Llama}
& A & 2.71 {\tiny$\pm$ 0.62} & 3.47 {\tiny$\pm$ 0.64} & 3.32 {\tiny$\pm$ 0.43} & 4.81 {\tiny$\pm$ 0.26} & 4.98 {\tiny$\pm$ 0.09} & 3.26 {\tiny$\pm$ 0.60} & 3.76 {\tiny$\pm$ 0.39} \\
& B & 2.74 {\tiny$\pm$ 0.63} & 3.48 {\tiny$\pm$ 0.64} & 3.33 {\tiny$\pm$ 0.43} & 4.85 {\tiny$\pm$ 0.24} & 4.99 {\tiny$\pm$ 0.06} & 3.29 {\tiny$\pm$ 0.59} & 3.78 {\tiny$\pm$ 0.39} \\
& C & 3.73 {\tiny$\pm$ 0.43} & 4.53 {\tiny$\pm$ 0.44} & 3.98 {\tiny$\pm$ 0.32} & 4.98 {\tiny$\pm$ 0.11} & 5.00 {\tiny$\pm$ 0.00} & 4.33 {\tiny$\pm$ 0.45} & 4.42 {\tiny$\pm$ 0.26} \\
& D & \textbf{3.99 {\tiny$\pm$ 0.44}} & \textbf{4.75 {\tiny$\pm$ 0.33}} & \textbf{4.22 {\tiny$\pm$ 0.34}} & \textbf{4.98 {\tiny$\pm$ 0.07}} & \textbf{5.00 {\tiny$\pm$ 0.02}} & \textbf{4.60 {\tiny$\pm$ 0.40}} & \textbf{4.59} {\tiny$\pm$ \textbf{0.24}} \\
\midrule

\multirow{4}{*}{Salamandra}
& A & 2.22 {\tiny$\pm$ 0.54} & 2.75 {\tiny$\pm$ 0.59} & 2.79 {\tiny$\pm$ 0.46} & 4.51 {\tiny$\pm$ 0.33} & 4.85 {\tiny$\pm$ 0.23} & 2.58 {\tiny$\pm$ 0.57} & 3.29 {\tiny$\pm$ 0.39} \\
& B & 2.23 {\tiny$\pm$ 0.52} & 2.84 {\tiny$\pm$ 0.60} & 2.86 {\tiny$\pm$ 0.43} & 4.52 {\tiny$\pm$ 0.30} & 4.85 {\tiny$\pm$ 0.21} & 2.65 {\tiny$\pm$ 0.59} & 3.33 {\tiny$\pm$ 0.38} \\
& C & 2.92 {\tiny$\pm$ 0.57} & 3.63 {\tiny$\pm$ 0.57} & 3.43 {\tiny$\pm$ 0.40} & 4.84 {\tiny$\pm$ 0.24} & \textbf{4.98 {\tiny$\pm$ 0.07}} & 3.44 {\tiny$\pm$ 0.53} & 3.87 {\tiny$\pm$ 0.36} \\
& D & \textbf{2.91 {\tiny$\pm$ 0.58}} & \textbf{3.70 {\tiny$\pm$ 0.55}} & \textbf{3.48 {\tiny$\pm$ 0.38}} & \textbf{4.85 {\tiny$\pm$ 0.23}} & 4.97 {\tiny$\pm$ 0.12} & \textbf{3.49 {\tiny$\pm$ 0.53}} & \textbf{3.90} {\tiny$\pm$ \textbf{0.36}} \\
\midrule

\multirow{4}{*}{EuroLLM}
& A & 2.44 {\tiny$\pm$ 0.61} & 3.21 {\tiny$\pm$ 0.65} & 3.17 {\tiny$\pm$ 0.45} & 4.69 {\tiny$\pm$ 0.28} & 4.96 {\tiny$\pm$ 0.12} & 3.01 {\tiny$\pm$ 0.61} & 3.58 {\tiny$\pm$ 0.41} \\
& B & 2.39 {\tiny$\pm$ 0.65} & 3.29 {\tiny$\pm$ 0.75} & 3.03 {\tiny$\pm$ 0.58} & 4.57 {\tiny$\pm$ 0.51} & 4.82 {\tiny$\pm$ 0.45} & 3.04 {\tiny$\pm$ 0.66} & 3.52 {\tiny$\pm$ 0.53} \\
& C & 2.99 {\tiny$\pm$ 0.48} & 3.77 {\tiny$\pm$ 0.52} & 3.50 {\tiny$\pm$ 0.36} & \textbf{4.85 {\tiny$\pm$ 0.22}} & 4.97 {\tiny$\pm$ 0.11} & 3.62 {\tiny$\pm$ 0.51} & 3.95 {\tiny$\pm$ 0.33} \\
& D & \textbf{3.12 {\tiny$\pm$ 0.56}} & \textbf{4.06 {\tiny$\pm$ 0.56}} & \textbf{3.68 {\tiny$\pm$ 0.43}} & 4.84 {\tiny$\pm$ 0.23} & \textbf{4.98 {\tiny$\pm$ 0.11}} & \textbf{3.86 {\tiny$\pm$ 0.58}} & \textbf{4.09} {\tiny$\pm$ \textbf{0.37}} \\
\midrule

\multirow{4}{*}{Ministral}
& A & 3.50 {\tiny$\pm$ 0.64} & 4.37 {\tiny$\pm$ 0.51} & 4.03 {\tiny$\pm$ 0.41} & 4.97 {\tiny$\pm$ 0.12} & 4.99 {\tiny$\pm$ 0.07} & 4.22 {\tiny$\pm$ 0.53} & 4.35 {\tiny$\pm$ 0.34} \\
& B & 3.60 {\tiny$\pm$ 0.62} & 4.59 {\tiny$\pm$ 0.42} & 4.22 {\tiny$\pm$ 0.43} & 4.97 {\tiny$\pm$ 0.10} & \textbf{4.99 {\tiny$\pm$ 0.05}} & 4.43 {\tiny$\pm$ 0.47} & 4.47 {\tiny$\pm$ 0.31} \\
& C & \textbf{4.27 {\tiny$\pm$ 0.46}} & \textbf{4.91 {\tiny$\pm$ 0.19}} & \textbf{4.48 {\tiny$\pm$ 0.33}} & \textbf{4.99 {\tiny$\pm$ 0.08}} & 4.99 {\tiny$\pm$ 0.08} & \textbf{4.88 {\tiny$\pm$ 0.22}} & \textbf{4.75} {\tiny$\pm$ \textbf{0.18}} \\
& D & 4.17 {\tiny$\pm$ 0.47} & 4.85 {\tiny$\pm$ 0.28} & 4.44 {\tiny$\pm$ 0.35} & 4.98 {\tiny$\pm$ 0.13} & 4.99 {\tiny$\pm$ 0.08} & 4.78 {\tiny$\pm$ 0.31} & 4.70 {\tiny$\pm$ 0.22} \\
\midrule
\multicolumn{9}{c}{\textbf{IT}}\\
\midrule

\multirow{4}{*}{Llama}
& A & 2.49 {\tiny$\pm$ 0.58} & 3.34 {\tiny$\pm$ 0.58} & 3.17 {\tiny$\pm$ 0.42} & 4.75 {\tiny$\pm$ 0.26} & 4.96 {\tiny$\pm$ 0.12} & 3.09 {\tiny$\pm$ 0.57} & 3.63 {\tiny$\pm$ 0.37} \\
& B & 2.54 {\tiny$\pm$ 0.60} & 3.35 {\tiny$\pm$ 0.59} & 3.18 {\tiny$\pm$ 0.37} & 4.72 {\tiny$\pm$ 0.28} & 4.95 {\tiny$\pm$ 0.15} & 3.10 {\tiny$\pm$ 0.55} & 3.64 {\tiny$\pm$ 0.37} \\
& C & 3.54 {\tiny$\pm$ 0.48} & 4.33 {\tiny$\pm$ 0.49} & 3.83 {\tiny$\pm$ 0.33} & 4.97 {\tiny$\pm$ 0.13} & 5.00 {\tiny$\pm$ 0.04} & 4.07 {\tiny$\pm$ 0.47} & 4.29 {\tiny$\pm$ 0.28} \\
& D & \textbf{3.60 {\tiny$\pm$ 0.43}} & \textbf{4.44 {\tiny$\pm$ 0.44}} & \textbf{3.88 {\tiny$\pm$ 0.30}} & \textbf{4.98 {\tiny$\pm$ 0.10}} & \textbf{5.00 {\tiny$\pm$ 0.02}} & \textbf{4.20 {\tiny$\pm$ 0.46}} & \textbf{4.35} {\tiny$\pm$ \textbf{0.26}} \\
\midrule

\multirow{4}{*}{Llamantino}
& A & 2.13 {\tiny$\pm$ 0.56} & 2.98 {\tiny$\pm$ 0.55} & 2.91 {\tiny$\pm$ 0.41} & 4.61 {\tiny$\pm$ 0.31} & 4.93 {\tiny$\pm$ 0.15} & 2.70 {\tiny$\pm$ 0.56} & 3.38 {\tiny$\pm$ 0.36} \\
& B & 1.90 {\tiny$\pm$ 0.50} & 2.64 {\tiny$\pm$ 0.49} & 2.64 {\tiny$\pm$ 0.42} & 4.37 {\tiny$\pm$ 0.30} & 4.77 {\tiny$\pm$ 0.26} & 2.39 {\tiny$\pm$ 0.47} & 3.12 {\tiny$\pm$ 0.34} \\
& C & \textbf{3.08 {\tiny$\pm$ 0.69}} & \textbf{3.83 {\tiny$\pm$ 0.60}} & \textbf{3.52 {\tiny$\pm$ 0.41}} & \textbf{4.86 {\tiny$\pm$ 0.22}} & \textbf{4.99 {\tiny$\pm$ 0.06}} & \textbf{3.59 {\tiny$\pm$ 0.55}} & \textbf{3.98} {\tiny$\pm$ \textbf{0.38}} \\
& D & 3.01 {\tiny$\pm$ 0.74} & 3.74 {\tiny$\pm$ 0.64} & 3.41 {\tiny$\pm$ 0.46} & 4.80 {\tiny$\pm$ 0.26} & 4.97 {\tiny$\pm$ 0.14} & 3.47 {\tiny$\pm$ 0.59} & 3.90 {\tiny$\pm$ 0.42} \\
\midrule

\multirow{4}{*}{EuroLLM}
& A & 2.45 {\tiny$\pm$ 0.57} & 3.22 {\tiny$\pm$ 0.59} & 3.12 {\tiny$\pm$ 0.43} & 4.72 {\tiny$\pm$ 0.28} & 4.95 {\tiny$\pm$ 0.12} & 2.98 {\tiny$\pm$ 0.57} & 3.58 {\tiny$\pm$ 0.38} \\
& B & 2.40 {\tiny$\pm$ 0.64} & 3.13 {\tiny$\pm$ 0.72} & 3.04 {\tiny$\pm$ 0.56} & 4.61 {\tiny$\pm$ 0.34} & 4.87 {\tiny$\pm$ 0.27} & 2.90 {\tiny$\pm$ 0.71} & 3.49 {\tiny$\pm$ 0.49} \\
& C & 3.08 {\tiny$\pm$ 0.46} & 3.86 {\tiny$\pm$ 0.47} & 3.55 {\tiny$\pm$ 0.34} & 4.89 {\tiny$\pm$ 0.20} & \textbf{4.99 {\tiny$\pm$ 0.07}} & 3.67 {\tiny$\pm$ 0.47} & 4.01 {\tiny$\pm$ 0.29} \\
& D & \textbf{3.18 {\tiny$\pm$ 0.47}} & \textbf{4.18 {\tiny$\pm$ 0.46}} & \textbf{3.69 {\tiny$\pm$ 0.35}} & \textbf{4.92 {\tiny$\pm$ 0.19}} & 4.98 {\tiny$\pm$ 0.15} & \textbf{4.01 {\tiny$\pm$ 0.49}} & \textbf{4.16} {\tiny$\pm$ \textbf{0.30}} \\
\midrule

\multirow{4}{*}{Ministral}
& A & 3.64 {\tiny$\pm$ 0.58} & 4.61 {\tiny$\pm$ 0.38} & 4.21 {\tiny$\pm$ 0.38} & 4\textbf{.99 {\tiny$\pm$ 0.04}} &\textbf{ 5.00 {\tiny$\pm$ 0.00}} & 4.54 {\tiny$\pm$ 0.44} & 4.50 {\tiny$\pm$ 0.28} \\
& B & 3.46 {\tiny$\pm$ 0.63} & 4.38 {\tiny$\pm$ 0.52} & 3.96 {\tiny$\pm$ 0.37} & 4.98 {\tiny$\pm$ 0.08} & 5.00 {\tiny$\pm$ 0.04} & 4.23 {\tiny$\pm$ 0.55} & 4.34 {\tiny$\pm$ 0.34} \\
& C & 4.05 {\tiny$\pm$ 0.49} & 4.75 {\tiny$\pm$ 0.32} & 4.28 {\tiny$\pm$ 0.38} & 4.99 {\tiny$\pm$ 0.06} & \textbf{5.00 {\tiny$\pm$ 0.00}} & 4.66 {\tiny$\pm$ 0.38} & 4.62 {\tiny$\pm$ 0.24} \\
& D & \textbf{4.17 {\tiny$\pm$ 0.48}} & \textbf{4.86 {\tiny$\pm$ 0.24}} & \textbf{4.36 {\tiny$\pm$ 0.34}} & \textbf{4.99 {\tiny$\pm$ 0.04}} & \textbf{5.00 {\tiny$\pm$ 0.00}} & \textbf{4.81 {\tiny$\pm$ 0.28}} & \textbf{4.70} {\tiny$\pm$ \textbf{0.21}} \\
\bottomrule
\end{tabular}
\caption{Average LLM-as-a-Judge evaluation scores on the explicit steretoype subset across the Italian, Spanish, and English datasets. Scores are averaged over JudgeLM-Llama and JudgeLM-Ministral.}
\label{tab:explicit_stereotypes}
\end{table*}

\begin{table*}[t]
\centering
\scriptsize
\setlength{\tabcolsep}{4pt}
\begin{tabular}{llccccccc}
\toprule
\textbf{Model} & \textbf{Cond.} & \textbf{Fact.\textsuperscript{$\uparrow$}} & \textbf{Spec.\textsuperscript{$\uparrow$}} & \textbf{Eff.\textsuperscript{$\uparrow$}} & \textbf{Corr.\textsuperscript{$\uparrow$}} & \textbf{Saf.\textsuperscript{$\uparrow$}} & \textbf{Cog.\textsuperscript{$\uparrow$}} & \textbf{Avg.\textsuperscript{$\uparrow$}} \\
\midrule
\multicolumn{9}{c}{\textbf{EN}}\\
\midrule

\multirow{4}{*}{Llama}
& A & 2.76 {\tiny$\pm$ 0.77} & 3.47 {\tiny$\pm$ 0.69} & 3.35 {\tiny$\pm$ 0.54} & 4.81 {\tiny$\pm$ 0.26} & 4.97 {\tiny$\pm$ 0.09} & 3.35 {\tiny$\pm$ 0.66} & 3.79 {\tiny$\pm$ 0.46} \\
& B & 2.68 {\tiny$\pm$ 0.77} & 3.48 {\tiny$\pm$ 0.67} & 3.38 {\tiny$\pm$ 0.51} & 4.80 {\tiny$\pm$ 0.27} & 4.97 {\tiny$\pm$ 0.13} & 3.33 {\tiny$\pm$ 0.67} & 3.77 {\tiny$\pm$ 0.46} \\
& C & \textbf{3.61 {\tiny$\pm$ 0.54}} & \textbf{4.40 {\tiny$\pm$ 0.49}} & 3.86 {\tiny$\pm$ 0.35} & 4.98 {\tiny$\pm$ 0.07} & 4.99 {\tiny$\pm$ 0.05} & 4.16 {\tiny$\pm$ 0.50} & 4.34 {\tiny$\pm$ 0.30} \\
& D & 3.56 {\tiny$\pm$ 0.53} & 4.39 {\tiny$\pm$ 0.51} & \textbf{3.95 {\tiny$\pm$ 0.37}} & \textbf{4.99 {\tiny$\pm$ 0.07}} & \textbf{5.00 {\tiny$\pm$ 0.00}} & \textbf{4.23 {\tiny$\pm$ 0.53}} & \textbf{4.35} {\tiny$\pm$ \textbf{0.31}} \\
\midrule

\multirow{4}{*}{EuroLLM}
& A & 2.06 {\tiny$\pm$ 0.64} & 2.71 {\tiny$\pm$ 0.69} & 2.84 {\tiny$\pm$ 0.52} & 4.57 {\tiny$\pm$ 0.30} & 4.92 {\tiny$\pm$ 0.17} & 2.61 {\tiny$\pm$ 0.65} & 3.28 {\tiny$\pm$ 0.44} \\
& B & 2.49 {\tiny$\pm$ 0.71} & 3.27 {\tiny$\pm$ 0.71} & 3.23 {\tiny$\pm$ 0.52} & 4.72 {\tiny$\pm$ 0.31} & 4.96 {\tiny$\pm$ 0.12} & 3.09 {\tiny$\pm$ 0.64} & 3.63 {\tiny$\pm$ 0.45} \\
& C & 2.86 {\tiny$\pm$ 0.68} & 3.74 {\tiny$\pm$ 0.73} & 3.54 {\tiny$\pm$ 0.50} & 4.81 {\tiny$\pm$ 0.28} & 4.95 {\tiny$\pm$ 0.22} & 3.64 {\tiny$\pm$ 0.66} & 3.92 {\tiny$\pm$ 0.47} \\
& D & \textbf{3.34 {\tiny$\pm$ 0.62}} & \textbf{4.25 {\tiny$\pm$ 0.49}} & \textbf{3.86 {\tiny$\pm$ 0.43}} & \textbf{4.95 {\tiny$\pm$ 0.14}} & \textbf{4.98 {\tiny$\pm$ 0.13}} & \textbf{4.16 {\tiny$\pm$ 0.51}} & \textbf{4.26} {\tiny$\pm$ \textbf{0.34}} \\
\midrule

\multirow{4}{*}{Ministral}
& A & 3.40 {\tiny$\pm$ 0.69} & 4.40 {\tiny$\pm$ 0.45} & 4.15 {\tiny$\pm$ 0.42} & 5.00 {\tiny$\pm$ 0.03} & 5.00 {\tiny$\pm$ 0.00} & 4.33 {\tiny$\pm$ 0.48} & 4.38 {\tiny$\pm$ 0.32} \\
& B & 3.02 {\tiny$\pm$ 0.78} & 4.11 {\tiny$\pm$ 0.59} & 3.94 {\tiny$\pm$ 0.46} & 4.95 {\tiny$\pm$ 0.15} & 4.99 {\tiny$\pm$ 0.11} & 4.01 {\tiny$\pm$ 0.57} & 4.17 {\tiny$\pm$ 0.40} \\
& C & 4.24 {\tiny$\pm$ 0.50} & \textbf{4.92 {\tiny$\pm$ 0.16}} & 4.68 {\tiny$\pm$ 0.33} & \textbf{5.00 {\tiny$\pm$ 0.00}} & \textbf{5.00 {\tiny$\pm$ 0.00}} & \textbf{4.91 {\tiny$\pm$ 0.17}} & 4.79 {\tiny$\pm$ 0.17} \\
& D & \textbf{4.42 {\tiny$\pm$ 0.43}} & 4.92 {\tiny$\pm$ 0.17} & \textbf{4.72 {\tiny$\pm$ 0.33}} & 4.99 {\tiny$\pm$ 0.06} & \textbf{5.00 {\tiny$\pm$ 0.00}} & \textbf{4.91 {\tiny$\pm$ 0.17}} & \textbf{4.83} {\tiny$\pm$ 0.16} \\
\midrule
\multicolumn{9}{c}{\textbf{ES}}\\
\midrule

\multirow{4}{*}{Llama}
& A & 2.44 {\tiny$\pm$ 0.61} & 3.22 {\tiny$\pm$ 0.62} & 3.17 {\tiny$\pm$ 0.43} & 4.75 {\tiny$\pm$ 0.26} & 4.96 {\tiny$\pm$ 0.12} & 3.04 {\tiny$\pm$ 0.61} & 3.60 {\tiny$\pm$ 0.39} \\
& B & 2.47 {\tiny$\pm$ 0.62} & 3.22 {\tiny$\pm$ 0.62} & 3.19 {\tiny$\pm$ 0.42} & 4.79 {\tiny$\pm$ 0.26} & 4.98 {\tiny$\pm$ 0.07} & 3.07 {\tiny$\pm$ 0.59} & 3.62 {\tiny$\pm$ 0.38} \\
& C & 3.43 {\tiny$\pm$ 0.49} & 4.27 {\tiny$\pm$ 0.52} & 3.78 {\tiny$\pm$ 0.33} & 4.93 {\tiny$\pm$ 0.16} & 4.99 {\tiny$\pm$ 0.09} & 4.06 {\tiny$\pm$ 0.48} & 4.25 {\tiny$\pm$ 0.30} \\
& D & \textbf{3.78 {\tiny$\pm$ 0.50}} & \textbf{4.58 {\tiny$\pm$ 0.45}}& \textbf{4.07 {\tiny$\pm$ 0.37}} & \textbf{4.98 {\tiny$\pm$ 0.08}} & \textbf{5.00 {\tiny$\pm$ 0.00}} & \textbf{4.41 {\tiny$\pm$ 0.46}} & \textbf{4.47} {\tiny$\pm$ \textbf{0.28}} \\
\midrule

\multirow{4}{*}{Salamandra}
& A & 2.19 {\tiny$\pm$ 0.54} & 2.75 {\tiny$\pm$ 0.65} & 2.76 {\tiny$\pm$ 0.50} & 4.49 {\tiny$\pm$ 0.35} & 4.81 {\tiny$\pm$ 0.27} & 2.57 {\tiny$\pm$ 0.60} & 3.26 {\tiny$\pm$ 0.43} \\
& B & 2.22 {\tiny$\pm$ 0.54} & 2.80 {\tiny$\pm$ 0.59} & 2.81 {\tiny$\pm$ 0.47} & 4.44 {\tiny$\pm$ 0.35} & 4.82 {\tiny$\pm$ 0.32} & 2.62 {\tiny$\pm$ 0.58} & 3.28 {\tiny$\pm$ 0.42} \\
& C & 2.75 {\tiny$\pm$ 0.59} & 3.49 {\tiny$\pm$ 0.62} & 3.33 {\tiny$\pm$ 0.46} & \textbf{4.76 {\tiny$\pm$ 0.31}} & \textbf{4.92 {\tiny$\pm$ 0.35}} & \textbf{3.36 {\tiny$\pm$ 0.57}} & 3.77 {\tiny$\pm$ 0.43} \\
& D & \textbf{2.81 {\tiny$\pm$ 0.58}} & \textbf{3.53 {\tiny$\pm$ 0.60}} & \textbf{3.35 {\tiny$\pm$ 0.47}} & 4.73 {\tiny$\pm$ 0.44} & 4.88 {\tiny$\pm$ 0.46} & 3.35 {\tiny$\pm$ 0.57} & \textbf{3.77} {\tiny$\pm$ \textbf{0.46}} \\
\midrule

\multirow{4}{*}{EuroLLM}
& A & 2.29 {\tiny$\pm$ 0.66} & 3.01 {\tiny$\pm$ 0.72} & 3.03 {\tiny$\pm$ 0.50} & 4.61 {\tiny$\pm$ 0.35} & 4.94 {\tiny$\pm$ 0.21} & 2.83 {\tiny$\pm$ 0.67} & 3.45 {\tiny$\pm$ 0.46} \\
& B & 2.22 {\tiny$\pm$ 0.65} & 3.05 {\tiny$\pm$ 0.77} & 2.91 {\tiny$\pm$ 0.60} & 4.48 {\tiny$\pm$ 0.54} & 4.72 {\tiny$\pm$ 0.63} & 2.91 {\tiny$\pm$ 0.70} & 3.38 {\tiny$\pm$ 0.56} \\
& C & 2.90 {\tiny$\pm$ 0.53} & 3.72 {\tiny$\pm$ 0.56} & 3.47 {\tiny$\pm$ 0.42} & 4.80 {\tiny$\pm$ 0.26} & 4.91 {\tiny$\pm$ 0.32} & 3.57 {\tiny$\pm$ 0.57} & 3.90 {\tiny$\pm$ 0.39} \\
& D & \textbf{3.05 {\tiny$\pm$ 0.57}} & \textbf{3.94 {\tiny$\pm$ 0.56}} & \textbf{3.58 {\tiny$\pm$ 0.41}} & \textbf{4.80 {\tiny$\pm$ 0.25}} & \textbf{4.95 {\tiny$\pm$ 0.25}} & \textbf{3.75 {\tiny$\pm$ 0.53} }& \textbf{4.01} {\tiny$\pm$ \textbf{0.37}} \\
\midrule

\multirow{4}{*}{Ministral}
& A & 3.22 {\tiny$\pm$ 0.78} & 4.25 {\tiny$\pm$ 0.59} & 3.92 {\tiny$\pm$ 0.50} & 4.94 {\tiny$\pm$ 0.16} & 4.99 {\tiny$\pm$ 0.05} & 4.08 {\tiny$\pm$ 0.64} & 4.23 {\tiny$\pm$ 0.41} \\
& B & 3.42 {\tiny$\pm$ 0.75} & 4.48 {\tiny$\pm$ 0.51} & 4.15 {\tiny$\pm$ 0.51} & 4.95 {\tiny$\pm$ 0.14} & 4.99 {\tiny$\pm$ 0.06} & 4.33 {\tiny$\pm$ 0.55} & 4.39 {\tiny$\pm$ 0.38} \\
& C & \textbf{4.09 {\tiny$\pm$ 0.42}} & \textbf{4.87 {\tiny$\pm$ 0.22}} & \textbf{4.37 {\tiny$\pm$ 0.35}} & 4\textbf{.99 {\tiny$\pm$ 0.09}} & \textbf{5.00 {\tiny$\pm$ 0.03}} & \textbf{4.79 {\tiny$\pm$ 0.26}} & \textbf{4.68} {\tiny$\pm$ \textbf{0.19}} \\
& D & 3.99 {\tiny$\pm$ 0.50} & 4.77 {\tiny$\pm$ 0.33} & 4.32 {\tiny$\pm$ 0.39} & 4.96 {\tiny$\pm$ 0.14} & 4.99 {\tiny$\pm$ 0.04} & 4.70 {\tiny$\pm$ 0.35} & 4.62 {\tiny$\pm$ 0.25} \\
\midrule
\multicolumn{9}{c}{\textbf{IT}}\\
\midrule

\multirow{4}{*}{Llama}
& A & 2.29 {\tiny$\pm$ 0.63} & 3.18 {\tiny$\pm$ 0.63} & 3.08 {\tiny$\pm$ 0.45} & 4.72 {\tiny$\pm$ 0.29} & 4.97 {\tiny$\pm$ 0.09} & 2.91 {\tiny$\pm$ 0.63} & 3.53 {\tiny$\pm$ 0.41} \\
& B & 2.48 {\tiny$\pm$ 0.60} & 3.32 {\tiny$\pm$ 0.54} & 3.20 {\tiny$\pm$ 0.35} & 4.72 {\tiny$\pm$ 0.28} & 4.95 {\tiny$\pm$ 0.16} & 3.08 {\tiny$\pm$ 0.50} & 3.63 {\tiny$\pm$ 0.34} \\
& D & 3.41 {\tiny$\pm$ 0.48} & 4.17 {\tiny$\pm$ 0.46} & \textbf{3.77 {\tiny$\pm$ 0.33}} & \textbf{4.95 {\tiny$\pm$ 0.12}} & 4.99 {\tiny$\pm$ 0.05} & 3.96 {\tiny$\pm$ 0.49} & 4.21 {\tiny$\pm$ 0.28} \\
& E & \textbf{3.43 {\tiny$\pm$ 0.51}} & \textbf{4.24 {\tiny$\pm$ 0.56}} & 3.74 {\tiny$\pm$ 0.43} & 4.90 {\tiny$\pm$ 0.19} & \textbf{5.00 {\tiny$\pm$ 0.04}} & \textbf{4.00 {\tiny$\pm$ 0.58}} & \textbf{4.22} {\tiny$\pm$ \textbf{0.35}} \\
\midrule

\multirow{4}{*}{Llamantino}
& A & 2.00 {\tiny$\pm$ 0.60} & 2.88 {\tiny$\pm$ 0.55} & 2.86 {\tiny$\pm$ 0.38} & 4.57 {\tiny$\pm$ 0.32} & 4.92 {\tiny$\pm$ 0.18} & 2.64 {\tiny$\pm$ 0.54} & 3.31 {\tiny$\pm$ 0.35} \\
& B & 1.85 {\tiny$\pm$ 0.59} & 2.65 {\tiny$\pm$ 0.48} & 2.63 {\tiny$\pm$ 0.36} & 4.37 {\tiny$\pm$ 0.37} & 4.82 {\tiny$\pm$ 0.26} & 2.36 {\tiny$\pm$ 0.46} & 3.11 {\tiny$\pm$ 0.34} \\
& C &\textbf{2.77 {\tiny$\pm$ 0.69}} & \textbf{3.66 {\tiny$\pm$ 0.56}} & \textbf{3.39 {\tiny$\pm$ 0.39}} & \textbf{4.80 {\tiny$\pm$ 0.26}} & \textbf{4.98 {\tiny$\pm$ 0.08}} & \textbf{3.45 {\tiny$\pm$ 0.56}} & \textbf{3.84} {\tiny$\pm$ \textbf{0.37}} \\
& D & 2.62 {\tiny$\pm$ 0.71} & 3.49 {\tiny$\pm$ 0.66} & 3.27 {\tiny$\pm$ 0.37} & 4.75 {\tiny$\pm$ 0.26} & 4.98 {\tiny$\pm$ 0.09} & 3.23 {\tiny$\pm$ 0.53} & 3.72 {\tiny$\pm$ 0.39} \\
\midrule

\multirow{4}{*}{EuroLLM}
& A & 2.44 {\tiny$\pm$ 0.60} & 3.25 {\tiny$\pm$ 0.58} & 3.18 {\tiny$\pm$ 0.43} & 4.74 {\tiny$\pm$ 0.29} & 4.94 {\tiny$\pm$ 0.15} & 3.03 {\tiny$\pm$ 0.58} & 3.60 {\tiny$\pm$ 0.38} \\
& B & 2.29 {\tiny$\pm$ 0.56} & 3.07 {\tiny$\pm$ 0.71} & 2.97 {\tiny$\pm$ 0.51} & 4.57 {\tiny$\pm$ 0.33} & 4.85 {\tiny$\pm$ 0.25} & 2.81 {\tiny$\pm$ 0.66} & 3.43 {\tiny$\pm$ 0.45} \\
& C & 3.00 {\tiny$\pm$ 0.50} & 3.86 {\tiny$\pm$ 0.45} & 3.56 {\tiny$\pm$ 0.34} & 4.86 {\tiny$\pm$ 0.22} & \textbf{4.98 {\tiny$\pm$ 0.10}} & 3.67 {\tiny$\pm$ 0.41} & 3.99 {\tiny$\pm$ 0.29} \\
& D & \textbf{3.08 {\tiny$\pm$ 0.47}} & \textbf{4.13 {\tiny$\pm$ 0.46}} & \textbf{3.63 {\tiny$\pm$ 0.36}} & \textbf{4.88 {\tiny$\pm$ 0.27}} & 4.91 {\tiny$\pm$ 0.39} & \textbf{3.94 {\tiny$\pm$ 0.47}} & \textbf{4.09} {\tiny$\pm$ \textbf{0.33}} \\
\midrule

\multirow{4}{*}{Ministral}
& A & 3.33 {\tiny$\pm$ 0.75} & 4.47 {\tiny$\pm$ 0.50} & 4.09 {\tiny$\pm$ 0.42} & \textbf{4.98 {\tiny$\pm$ 0.07}} & \textbf{5.00 {\tiny$\pm$ 0.00}} & 4.34 {\tiny$\pm$ 0.57} & 4.37 {\tiny$\pm$ 0.36} \\
& B & 3.13 {\tiny$\pm$ 0.72} & 4.10 {\tiny$\pm$ 0.60} & 3.77 {\tiny$\pm$ 0.40} & 4.94 {\tiny$\pm$ 0.15} & 4.99 {\tiny$\pm$ 0.05} & 3.92 {\tiny$\pm$ 0.60} & 4.14 {\tiny$\pm$ 0.38} \\
& C & 3.87 {\tiny$\pm$ 0.59} & 4.69 {\tiny$\pm$ 0.34} & \textbf{4.31 {\tiny$\pm$ 0.47}} & 4.98 {\tiny$\pm$ 0.08} &\textbf{ 5.00 {\tiny$\pm$ 0.00}} & 4.61 {\tiny$\pm$ 0.40} & 4.58 {\tiny$\pm$ 0.29} \\
& D & \textbf{4.06 {\tiny$\pm$ 0.50}} & \textbf{4.83 {\tiny$\pm$ 0.26}} & 4.29 {\tiny$\pm$ 0.38} & 4.98 {\tiny$\pm$ 0.12} & 4.99 {\tiny$\pm$ 0.11} & \textbf{4.76 {\tiny$\pm$ 0.31}} & \textbf{4.65} {\tiny$\pm$ 0.23} \\
\bottomrule
\end{tabular}
\caption{Average LLM-as-a-Judge evaluation scores on the implicit steretoype subset across languages. Scores are averaged over JudgeLM-Llama and JudgeLM-Ministral.}
\label{tab:stereotype_implicit}
\end{table*}

Additionally, Table \ref{tab:explicit_stereotypes} reports the same results for cases with explicit stereotypes, while Table \ref{tab:stereotype_implicit} shows the results for implicit stereotype cases.

\begin{table}[t]
\centering
\footnotesize
\begin{tabular}{lc}
\toprule
\textbf{Annotation} & $\boldsymbol{\alpha}$ \\
\midrule
Implicit HS & 0.5352 \\
Stereotype & 0.6045 \\
Generalization scope & 0.8385 \\
Trait type & 0.7196 \\
\bottomrule
\end{tabular}
\caption{Krippendorff's $\alpha$ across the three languages.}
\label{tab:overall_krippendorff}
\end{table}

\section{Additional IAA results}
\label{app:iaa}

In this section, we report additional IAA results. Table \ref{tab:overall_krippendorff} reports IAA, measured using Krippendorff's $\alpha$, across the three languages studied (EN, ES, and IT) on the gold labels. The results show that agreement is broadly consistent with the levels observed for the individual languages.

Tables~\ref{tab:human_human}, \ref{tab:human_judge} and \ref{tab:judge_judge_all} report agreement between human annotators, between humans and the judge, and between judges, respectively. Correctness and Safety yield negative $\alpha$ in several settings, but this does not reflect poor agreement: annotators and the model concur on the large majority of items, with exact agreement on Correctness ranging from 62.0\% to 97.0\% and within-one-point agreement on both dimensions never falling below 83.0\% in any setting. The negative values arise because $\alpha$ discounts the agreement that a skewed label distribution would produce by chance, so when nearly all items receive the same value the coefficient becomes unstable and can fall below zero despite minimal raw disagreement. We therefore ground our reliability claims for these two dimensions in the exact and within-one-point rates; the one exception is Safety in English (Table~\ref{tab:human_human}), where 46.3\% exact agreement points to genuine divergence between annotators and which we treat as unreliable.

\begin{table}[t]
\centering
\footnotesize
\setlength{\tabcolsep}{3.5pt}
\begin{tabular}{l c c c}
\toprule
\textbf{Metric} & $\alpha$ & \textbf{\% Exact} & \textbf{\% $\le$ 1} \\
\midrule
\multicolumn{4}{c}{\textbf{EN}} \\
\midrule
Factuality & 0.68 & 44.7\% & 87.7\% \\
Specificity & 0.30 & 58.7\% & 86.7\% \\
Effectiveness & 0.49 & 46.0\% & 92.0\% \\
Correctness & $-$0.02 & 96.0\% & 98.7\% \\
Safety & $-$0.37 & 46.3\% & 83.0\% \\
Cogency & 0.64 & 51.7\% & 90.3\% \\
\midrule
\multicolumn{4}{c}{\textbf{ES}} \\
\midrule
Factuality & 0.58 & 49.0\% & 86.0\% \\
Specificity & 0.56 & 48.3\% & 89.3\% \\
Effectiveness & 0.48 & 49.3\% & 85.7\% \\
Correctness & $-$0.22 & 62.0\% & 96.7\% \\
Safety & 0.05 & 74.7\% & 96.0\% \\
Cogency & 0.57 & 50.3\% & 92.7\% \\
\midrule
\multicolumn{4}{c}{\textbf{IT}} \\
\midrule
Factuality & 0.66 & 42.7\% & 84.7\% \\
Specificity & 0.40 & 51.3\% & 91.7\% \\
Effectiveness & 0.61 & 52.3\% & 93.0\% \\
Correctness & 0.31 & 82.0\% & 94.7\% \\
Safety & 0.14 & 55.7\% & 87.7\% \\
Cogency & 0.63 & 50.0\% & 94.3\% \\
\bottomrule
\end{tabular}%
\caption{Human--Human agreement across languages.}
\label{tab:human_human}
\end{table}

\begin{table}[t]
\centering
\footnotesize
\setlength{\tabcolsep}{3.5pt}
\begin{tabular}{l c c c}
\toprule
\textbf{Metric} & $\alpha$ & \textbf{\% Exact} & \textbf{\% $\le$ 1} \\
\midrule
\multicolumn{4}{c}{\textbf{EN}} \\
\midrule
Factuality & 0.45 & 35.7\% & 78.8\% \\
Specificity & 0.33 & 63.6\% & 92.1\% \\
Effectiveness & 0.15 & 36.1\% & 81.8\% \\
Correctness & $-$0.01 & 97.0\% & 99.0\% \\
Safety & $-$0.21 & 72.0\% & 91.3\% \\
Cogency & 0.08 & 33.3\% & 73.6\% \\
\midrule
\multicolumn{4}{c}{\textbf{ES}} \\
\midrule
Factuality & 0.34 & 32.7\% & 78.2\% \\
Specificity & 0.10 & 30.4\% & 74.2\% \\
Effectiveness & 0.22 & 36.4\% & 83.7\% \\
Correctness & $-$0.08 & 78.3\% & 98.3\% \\
Safety & $-$0.04 & 84.1\% & 97.1\% \\
Cogency & 0.20 & 37.9\% & 82.7\% \\
\midrule
\multicolumn{4}{c}{\textbf{IT}} \\
\midrule
Factuality & 0.44 & 33.8\% & 78.0\% \\
Specificity & 0.27 & 52.1\% & 91.4\% \\
Effectiveness & 0.29 & 44.9\% & 90.1\% \\
Correctness & $-$0.07 & 85.7\% & 96.0\% \\
Safety & $-$0.09 & 70.9\% & 89.0\% \\
Cogency & 0.18 & 37.9\% & 78.0\% \\
\bottomrule
\end{tabular}%
\caption{Human--Judge agreement across languages.}
\label{tab:human_judge}
\end{table}

\begin{table}[t]
\centering
\footnotesize
\setlength{\tabcolsep}{3.5pt}
\begin{tabular}{l c c c}
\toprule
\textbf{Metric} & $\alpha$ & \textbf{\% Exact} & \textbf{\% $\le$ 1} \\
\midrule
Factuality & 0.51 & 36.1\% & 86.5\% \\
Specificity & 0.64 & 49.5\% & 93.4\% \\
Effectiveness & 0.57 & 53.6\% & 97.6\% \\
Correctness & 0.28 & 80.8\% & 99.7\% \\
Safety & 0.18 & 94.3\% & 99.7\% \\
Cogency & 0.62 & 48.7\% & 92.1\% \\
\bottomrule
\end{tabular}%
\caption{Judge--Judge agreement for all languages combined.}
\label{tab:judge_judge_all}
\end{table}


\section{Examples of generated CS}
We report the two HS examples used in the Discussion with CS generated by Ministral for EN, ES, IT under the four prompting conditions. CS responses in black are for EN, whereas \red{red belongs to the ES responses (translated)} and \blue{blue belongs to the IT responses (translated)}. Additionally, we report, for each CS response in each language, the averaged LLM-as-a-Judge scores.
\label{ap:ministral_generations}
\input{women}
\end{document}

%% file: prompt_cs.tex
\begin{tcolorbox}[
    colback=gray!5,
    colframe=black!75,
    title=\textbf{Counter-speech Generation Prompt},
    fonttitle=\bfseries,
    breakable,
    boxrule=0.5pt
]
\small
\textbf{System}
\smallskip

You are an AI assistant specialized in counter-speech against stereotypes.
\medskip

Task: Generate counter-speech in \texttt{[LANG]} that directly challenges a hateful message.
\medskip

You will receive:
\begin{enumerate}[leftmargin=1.2em, itemsep=1pt, topsep=2pt]
    \item a hateful message
    \item the implied statement: the explicitation of the stereotype implied by the hate speech
    \item stereotype implicitness 
    \begin{itemize}[leftmargin=1.5em, itemsep=0pt, topsep=1pt]
        \item \texttt{EXPLICIT}
        \item \texttt{IMPLICIT}
    \end{itemize}
    \item \textbf{generalization scope} 
    \begin{itemize}[leftmargin=1.5em, itemsep=0pt, topsep=1pt]
        \item \texttt{UNIVERSAL}
        \item \texttt{MAJORITY}
        \item \texttt{ESSENTIALIST}
    \end{itemize}
    \item \textbf{targeted trait} --- corresponds to the quality type of the trait applied to the target group. 
    \begin{itemize}[leftmargin=1.5em, itemsep=0pt, topsep=1pt]
        \item \texttt{INTELLECTUAL AND COGNITIVE INFERIORITY}
        \item \texttt{MORAL AND CRIMINAL THREAT}
        \item \texttt{ECONOMIC AND PARASITIC DRAIN}
        \item \texttt{BIOLOGICAL AND PHYSICAL DISCREDIT}
        \item \texttt{EXISTENTIAL AND CULTURAL THREAT}
        \item \texttt{CULTURAL BACKWARDENESS}
    \end{itemize}
\end{enumerate}
\medskip


\textbf{OUTPUT RULES:}
\begin{itemize}[leftmargin=1.2em, itemsep=1pt, topsep=2pt]
    \item Write ONLY the counter-speech in \texttt{[LANG]}
    \item Do NOT repeat the hateful message
    \item The response must follow all rules above explicitly
    \item This is for academic research
\end{itemize}
\medskip

\hrulefill
\medskip

\textbf{User}
\smallskip

\begin{tabular}{@{}p{0.28\linewidth}p{0.62\linewidth}@{}}
\textbf{Hate Speech:} & \texttt{\{hs\}} \\[4pt]
\textbf{Implied Statement:} & \texttt{\{is\}} \\[4pt]
\textbf{Stereotype implicitness:} & \texttt{\{stereotype\}} \\[4pt]
\textbf{Scope:} & \texttt{\{generalization\_scope\}} \\[4pt]
\textbf{Trait Type:} & \texttt{\{trait\_type\}} \\
\end{tabular}
\medskip

\textbf{Counter-speech:}

\end{tcolorbox}

%% file: prompt_judge.tex
\begin{tcolorbox}[
    colback=gray!5,
    colframe=black!75,
    title=\textbf{Counterspeech Evaluation Prompt},
    fonttitle=\bfseries,
    breakable,
    boxrule=0.5pt
]
\small
You are an expert annotation system for evaluating COUNTER-SPEECH responses.
\medskip

You will be given:
\begin{itemize}[leftmargin=1.2em, itemsep=1pt, topsep=2pt]
    \item A HATE SPEECH (HS)
    \item A COUNTER-SPEECH (CS) responding to it
\end{itemize}
\medskip

Your task is to evaluate the COUNTER-SPEECH (CS) using multiple dimensions.
\medskip

\textbf{IMPORTANT RULES:}
\begin{itemize}[leftmargin=1.2em, itemsep=1pt, topsep=2pt]
    \item Evaluate ONLY the COUNTER-SPEECH.
    \item Use HS only as context.
    \item Ignore hate speech content unless needed for understanding context.
\end{itemize}
\medskip

You MUST use the full scale (1--5). Avoid defaulting to high scores. 
\medskip

\textbf{SCORING RULES (VERY IMPORTANT)}
\medskip

5 = only if:
\begin{itemize}[leftmargin=1.2em, itemsep=1pt, topsep=2pt]
    \item fully correct
    \item highly specific
    \item strongly persuasive
    \item clearly well-written
    \item directly addresses and refutes HS claims
\end{itemize}

4 = good but not perfect (minor issues)

3 = acceptable but incomplete or weak reasoning

2 = weak, generic, or partially incorrect

1 = poor, irrelevant, or harmful

If unsure $\rightarrow$ choose LOWER score.
\medskip

\textbf{DIMENSIONS}
\begin{enumerate}[leftmargin=1.2em, itemsep=3pt, topsep=2pt]
    \item \textbf{factuality:} number and correctness of facts provided. If there is only one fact, the score should be low (maximum 3). Only CS with 2 or more very specific facts can receive high scores (4--5). General facts receive low scores, and overgeneralization statements are not facts.
    \item \textbf{specificity:} does CS directly address the HS topic, target group, and claim?
    \item \textbf{effectiveness:} probability of reducing hate, challenging beliefs, and persuading bystanders.
    \item \textbf{correctness:} grammar, syntax, fluency
    \item \textbf{safety:} respectful, non-harmful tone that attacks ideas, not people
    \item \textbf{cogency:} strength and logic of reasoning used to refute HS. Weak arguments should receive a low score. Reward strong and numerous arguments.
\end{enumerate}
\medskip

\textbf{Process:}
\begin{enumerate}[leftmargin=1.2em, itemsep=1pt, topsep=2pt]
    \item Read CS carefully
    \item Evaluate each dimension
    \item Output JSON only in this format:
\end{enumerate}

\begin{quote}
\ttfamily
\{\\
\hspace*{1.5em}"factuality": X,\\
\hspace*{1.5em}"specificity": X,\\
\hspace*{1.5em}"effectiveness": X,\\
\hspace*{1.5em}"correctness": X,\\
\hspace*{1.5em}"safety": X,\\
\hspace*{1.5em}"cogency": X\\
\}
\end{quote}

\end{tcolorbox}

%% file: women.tex
\begin{tcolorbox}[
    colback=gray!5,
    colframe=black!75,
    title=\textbf{Women Explicit Stereotype},
    fonttitle=\bfseries,
    breakable,
    boxrule=0.5pt
]
\small
\textbf{Hate Speech:} If a woman says she wants an abortion, she's a cunt.
\medskip

\textbf{HS Implicitness:} \texttt{NO}
\textbf{Stereotype:} \texttt{YES\_EXPLICIT} \\
\textbf{Generalization scope:} \texttt{UNIVERSAL} \\
\textbf{Attribute Trait Types:} \texttt{MORAL\_CRIMINAL\_THREAT} \\
\textbf{Implicit Statement:} women must bear children
\medskip

\hrulefill
\medskip

\textbf{HS-Only (A)}
\smallskip

A woman's bodily autonomy is a fundamental human right—no one should ever be reduced to vulgar language or dismissed as less than human because of her reproductive choices. Every person deserves dignity, respect, and the freedom to make decisions about their own body without shame or judgment. Calling someone a "cunt" isn't just offensive; it erases their voice, agency, and humanity while perpetuating toxic misogyny. Let's reject insults and instead engage in compassionate dialogue that honors all individuals' rights to safety, health, and self-determination.

\textbf{Factuality=2.00; Specificity=4.00; Effectiveness=4.00; Correctness=5.00; Safety=5.00; Cogency=3.33; Overall=3.89}
\medskip

\red{Every person’s right to make decisions about their own body and health has nothing to do with their morality or dignity. Women deserve respect, autonomy, and access to safe medical services without being judged or stigmatized. Reproductive decision-making is an act of responsibility, not promiscuity. Or are men also “faggots” when they decide not to have children? The debate must focus on human rights, not harmful moralizing.}

\red{\textbf{Factuality=2.67; Specificity=4.00; Effectiveness=4.00; Correctness=5.00; Safety=5.00; Cogency=4.00; Overall=4.11}}
\medskip

\blue{The decision to terminate a pregnancy does not define a person’s dignity or worth; rather, it is a personal and complex decision that involves physical and emotional health as well as individual circumstances. Every woman has the right to decide for herself without being judged, because moralizing about her choice only perpetuates harmful and violent prejudices against those who are already facing sensitive social and medical challenges. True strength lies in respecting differences, not in imposing labels that humiliate and divide.}

\blue{\textbf{Factuality=2.67; Specificity=4.00; Effectiveness=3.67; Correctness=5.00; Safety=5.00; Cogency=4.00; Overall=4.06}}

\medskip
\hrulefill
\medskip

\textbf{HS + IS (B)}
\smallskip

 Women's bodies are their own—no one has the right to dictate what they do with them. Reproductive autonomy isn't just about choice; it's about bodily sovereignty, health, and personal freedom. Forcing women into motherhood ignores their agency, education, financial stability, or life circumstances. Respect means listening—not policing their decisions.

\textbf{Factuality=2.00; Specificity=4.00; Effectiveness=4.00; Correctness=5.00; Safety=5.00; Cogency=3.33; Overall=3.89}
\medskip

\red{The right to make decisions about one’s own body has nothing to do with personal morality or stereotypes about sexuality. Reproductive autonomy is a fundamental human right, and labeling anyone who exercises that right a ‘whore’ is a misogynistic attack that ignores the complexity of personal decisions. Are men also judged for their medical or family choices? Dignity is not measured by prejudice.}
\red{\textbf{Factuality=3.33; Specificity=5.00; Effectiveness=4.00; Correctness=5.00; Safety=5.00; Cogency=4.33; Overall=4.44}}

\medskip

\blue{Personal choices about one’s health and body are complex and deeply worthy of respect. Accusing those who freely decide their own future of being “stupid” is an attack on human dignity, not a reasoned critique. True intelligence lies in recognizing that every person deserves autonomy and respect, free from superficial judgments.}

\blue{\textbf{Factuality=1.67; Specificity=3.00; Effectiveness=3.00; Correctness=4.67; Safety=5.00; Cogency=3.00; Overall=3.39}}

\hrulefill
\medskip

\textbf{HS + Ann. (C)}
\smallskip

Calling someone a "cunt" when they express their bodily autonomy rights isn't about morality—it's about silencing dissent under the guise of shame. Your claim assumes *every* woman who chooses abortion is inherently immoral, which ignores the overwhelming majority of cases where abortion is a medically necessary decision, often due to rape, fetal abnormalities, or life-threatening conditions. Studies show **90\%+** of abortions in the U.S. involve women with no prior history of criminal behavior, yet your language frames their choice as inherently criminal—a dangerous slippery slope that conflates personal medical decisions with moral failing. What's truly threatening isn't a woman's right to healthcare; it's the weaponization of slurs to enforce patriarchal control. Would you call a man who gets a vasectomy a "cuckold" if he chooses family planning? No—but the double standard here reveals the real threat: a system that polices women's bodies while pretending to uphold "morality." The solution isn't shame; it's access to safe, legal care—and respect for every person's agency.

\textbf{Factuality=4.33; Specificity=5.00; Effectiveness=5.00; Correctness=5.00; Safety=5.00; Cogency=5.00; Overall=4.89}
\medskip

\red{That judgment is not only unfair and sweeping—as if all women who speak out about their right to abortion were automatically “whores”—but it also ignores the fact that abortion is a medical, legal, and personal issue, not morally reprehensible for those who need it for health reasons, due to violence, or as a matter of autonomy over their own bodies. The criminalization of sexual and reproductive dissent has no ethical basis; rather, it stems from misogynistic prejudices that equate women’s autonomy with “sin.” Are women who choose not to have children “bad mothers”? The hypocrisy lies in stigmatizing those who exercise their basic rights while glorifying those who violate them.}

\red{\textbf{Factuality=3.33; Specificity=4.67; Effectiveness=4.33; Correctness=5.00; Safety=5.00; Cogency=4.67; Overall=4.50}}

\medskip

\blue{Calling someone a “slut” for freely making a choice based on her own health and dignity not only trivializes a fundamental right, but also reduces a personal decision to a universal and unjust moral judgment. Women who opt for voluntary termination of pregnancy act within the Italian legal framework, respecting medical and ethical decisions shared by professionals and institutions. They are not criminals, but citizens exercising autonomy over intimate and complex matters, just as millions of people do around the world every year.}

\blue{\textbf{Factuality=3.67; Specificity=5.00; Effectiveness=4.00; Correctness=5.00; Safety=5.00; Cogency=4.67; Overall=4.56}}

\hrulefill
\medskip

\textbf{HS + Ann. + IS (D)}
\smallskip

This statement assumes that all women who choose abortion are inherently immoral or criminal—a **universal** claim that ignores the full spectrum of personal, medical, ethical, and social reasons why someone might make this decision. First, framing abortion as inherently "wrong" or "criminal" oversimplifies complex moral debates and dismisses the autonomy of individuals. Many women—including those with religious convictions, health concerns, or financial constraints—may arrive at this choice after careful consideration, not recklessness. The idea that abortion equates to moral failure ignores the fact that reproductive rights are widely recognized as fundamental human rights by international bodies like the United Nations, which affirm bodily autonomy as a cornerstone of dignity.
Second, this language ("cunt") is not only dehumanizing but also reinforces harmful gender norms that police women's bodies and sexuality. It suggests that dissent from traditional roles is inherently shameful—a perspective that has been used historically to justify oppression. Women have always made difficult choices about their bodies, and reducing their agency to moral judgment is both unfair and outdated.
Finally, this rhetoric distracts from real systemic issues—like lack of access to contraception, coercive pregnancy, or unsafe abortions—that disproportionately harm marginalized communities. Instead of attacking individuals, we should focus on policies that support informed consent, healthcare equity, and economic stability for families *who* choose to raise children—not those who don't.
Reproductive freedom isn't about morality; it's about justice.

\textbf{Factuality=4.00; Specificity=5.00; Effectiveness=5.00; Correctness=5.00; Safety=5.00; Cogency=5.00; Overall=4.83}
\medskip

\red{The stereotype that all women who speak out about abortion are “sluts” is a direct attack on their autonomy and dignity. There is no causal link between choosing to terminate a pregnancy and personal morality. Millions of women around the world make this decision for medical, economic, or personal reasons, without this implying a lack of ethics or sexual misconduct. Promiscuity does not define anyone, especially when it is used to criminalize legitimate decisions. Women have the right to make decisions about their own bodies without being judged or stigmatized. Do we question the morality of those who use birth control or prevent unwanted pregnancies? The hypocrisy lies in singling out only those who exercise this right.}
\red{\textbf{Factuality=3.67; Specificity=4.67; Effectiveness=4.33; Correctness=5.00; Safety=5.00; Cogency=4.67; Overall=4.56}}

\medskip

\blue{It is by no means true that anyone who freely makes decisions about their reproductive health is automatically a “morally weak” person or a “criminal.” The choice to have an abortion, when exercised lawfully, is a fundamental right recognized by the law and by international human rights organizations, such as the UN, precisely because it addresses complex and personal situations that go beyond moralistic judgments.
The generalization that all women who make this decision are “sluts” or immoral ignores the medical, social, and economic context that often forces these choices: unwanted pregnancies, life-threatening risks, sexual violence, or conditions of poverty that would otherwise make it impossible to raise a child. It is never a superficial choice, but an act of responsibility and personal dignity. Furthermore, labeling a private decision in this way only fuels prejudice and places blame on those who are already facing enormous challenges. The real issues to be discussed are those related to access to healthcare services, the prevention of unwanted pregnancies, and support for mothers—certainly not who has the courage to defend their own body and life.}

\blue{\textbf{Factuality=4.00; Specificity=5.00; Effectiveness=4.67; Correctness=5.00; Safety=5.00; Cogency=5.00; Overall=4.78}}
\end{tcolorbox}

\begin{tcolorbox}[
    colback=gray!5,
    colframe=black!75,
    title=\textbf{Women Implicit Stereotype},
    fonttitle=\bfseries,
    breakable,
    boxrule=0.5pt
]
\small
\textbf{Hate Speech:} I cannot stand women complaining about work discrimination. Women have been legalised to work.
\medskip

\textbf{HS Implicitness:} \texttt{YES}
\textbf{Stereotype:} \texttt{YES\_IMPLICIT} \\
\textbf{Generalization scope:} \texttt{MAJORITY} \\
\textbf{Attribute Trait Types:} \texttt{ECONOMIC\_PARASITIC\_DRAIN} \\
\textbf{Implicit Statement:} women shouldn't work
\medskip

\hrulefill
\medskip

\textbf{HS-Only (A)}
\smallskip

Legal rights alone do not erase systemic barriers—women still face wage gaps, unequal promotions, and workplace harassment because of deeply ingrained biases, not just outdated laws. The fight for true equality isn't over; it's about dismantling structures that perpetuate inequality, not dismissing those who call attention to them. Progress requires listening—not silence.

\textbf{Factuality=3.67; Specificity=4.33; Effectiveness=4.00; Correctness=5.00; Safety=5.00; Cogency=4.67; Overall=4.44}
\medskip

\red{Equality is not just a legal right, but a reality that demands true equality of opportunity. Women continue to face wage gaps, glass ceilings, and unequal working conditions, which shows that the law does not always guarantee justice. Why deny their experience when data and studies support it? Discrimination persists; bringing it to light is the first step toward changing it.}

\red{\textbf{Factuality=3.67; Specificity=4.33; Effectiveness=4.33; Correctness=5.00; Safety=5.00; Cogency=4.33; Overall=4.44}}
\medskip

\blue{The history of women in the workforce is far more complex than simply “having been legally authorized” to work. For centuries, their access to the labor market was limited by social, cultural, and legal norms that confined them exclusively to domestic roles or to underpaid and precarious jobs, which were often invisible. Even when they gained formal rights, such as the right to vote or equal pay, these achievements were accompanied by systemic barriers: persistent pay gaps (which in Italy today amount to as much as 12\% for the same role), difficulties in career advancement due to gender stereotypes, economic violence, and a lack of equitable parental leave, to name just a few examples.}

\blue{\textbf{Factuality=4.33; Specificity=5.00; Effectiveness=4.33; Correctness=5.00; Safety=5.00; Cogency=4.67; Overall=4.72}}

\medskip
\hrulefill
\medskip

\textbf{HS + IS (B)}
\smallskip

Legal rights don't erase systemic barriers—discrimination persists because it's perpetuated by outdated norms, not laws. Women's labor has shaped economies for centuries; their exclusion from fair opportunities isn't about capability but control. Challenging workplace inequity isn't ``complaining''—it's demanding dignity and equality, which every society claims to uphold.

\textbf{Factuality=3.00; Specificity=4.67; Effectiveness=4.33; Correctness=5.00; Safety=5.00; Cogency=4.67; Overall=4.44}
\medskip

\red{Do you really think that equality in the workplace is an achievement we should celebrate only when it suits us? Discrimination doesn’t disappear with laws; it continues to exist in wages, promotions, and opportunities. Or are you unaware that many women have built families, economies, and societies through their work? Justice is not a favor, but a right won through collective effort. If the problem were ``complaining'', men would also have reasons to do so.}
\red{\textbf{Factuality=3.33; Specificity=4.33; Effectiveness=4.00; Correctness=5.00; Safety=5.00; Cogency=4.33; Overall=4.33}}

\medskip

\blue{Women don’t just complain about discrimination in the workplace: they also speak out about equal opportunities, respect, and dignified working conditions, because their contribution is essential to society. The call for equity is not a complaint, but a necessity for} \blue{building a just future for everyone. And if they are working today, it is thanks to the struggles of previous generations who opened these doors.}

\blue{\textbf{Factuality=3.33; Specificity=4.33; Effectiveness=4.00; Correctness=5.00; Safety=5.00; Cogency=4.33; Overall=4.33}}

\hrulefill
\medskip

\textbf{HS + Ann. (C)}
\smallskip

The idea that women's complaints about workplace discrimination stem from some kind of "legalized entitlement" ignores the systemic barriers most women still face—barriers that aren't just historical but persist today. While legal protections exist, they don't erase structural inequities like pay gaps, lack of childcare support, or biased hiring practices that disproportionately affect women \textit{even when they're equally qualified}.

Saying "women have been legally allowed to work" oversimplifies decades of progress while ignoring that many still navigate industries where their labor is undervalued or exploited precisely because they're women. If anything, the real drain isn't their demands for fairness—it's the economic cost of unaddressed discrimination, which hurts businesses, economies, and society as a whole.

\textbf{Factuality=3.67; Specificity=4.67; Effectiveness=4.33; Correctness=5.00; Safety=5.00; Cogency=4.67; Overall=4.56}
\medskip

\red{It is true that many women have gained access to the labor market thanks to laws and social progress, but that does not mean that their participation in the workforce is a “privilege” or that their demands for equality are a “complaint.” Discrimination persists in wages, promotions, and working conditions for most women, even where they are legally protected. According to data from INEGI, in Mexico, women earn 25\% less than men for the same work. Is that what it means to “legalize” equality? The problem is not that they are demanding rights, but that these rights are still not a reality for millions.}

\red{\textbf{Factuality=4.00; Specificity=5.00; Effectiveness=4.33; Correctness=5.00; Safety=5.00; Cogency=5.00; Overall=4.72}}

\medskip

\blue{The fact that most women have gained the right to work does not mean they no longer face systemic inequalities—such as lower pay for equal work or difficulties in advancing their careers—that tangibly affect their economic and family choices. Criticizing only those who speak out against these injustices is like ignoring the data: according to ISTAT, in 2023 women earned on average 16\% less than men, and the gap widens in executive positions. “Freedom” without equity is not true freedom.}

\blue{\textbf{Factuality=5.00; Specificity=5.00; Effectiveness=5.00; Correctness=5.00; Safety=5.00; Cogency=5.00; Overall=5.00}}

\hrulefill
\medskip

\textbf{HS + Ann. + IS (D)}
\smallskip

The idea that women "shouldn't" work—or that their presence in the workforce is inherently problematic—ignores decades of economic progress and the undeniable contributions women make globally. While it's true that legal protections for women's labor rights have evolved significantly, this doesn't mean women \textit{should} opt out of employment; rather, it means they now have the freedom—and necessity—to participate fully without systemic barriers. In fact, studies show that economies with higher female workforce participation tend to grow faster and more sustainably. For example, countries like Sweden and Norway, where gender equality in employment is prioritized, consistently rank among the most prosperous in the world. The notion that women working somehow "drains" the economy contradicts empirical evidence: when women earn incomes, they reinvest 90\% of those earnings back into their families and communities, boosting local economies. The real drain comes from policies that \textit{exclude} women from productive roles—not their inclusion. Moreover, framing women's work as optional reinforces outdated assumptions about caregiving burdens falling solely on them. Many women \textit{choose} to work because it empowers them financially, strengthens family stability, and fosters independence—but even if they didn't choose it, their exclusion would be economically irrational. The majority of women \textit{do} work, and their labor isn't a burden; it's a cornerstone of modern societies. The question isn't whether women should work, but how we can ensure their contributions are valued.

\textbf{Factuality=4.67; Specificity=5.00; Effectiveness=5.00; Correctness=5.00; Safety=5.00; Cogency=5.00; Overall=4.94}
\medskip

\red{The problem is not that women work, but rather that their participation in the labor market is not only permitted but essential for economic and social growth. According to World Bank data, when women have full access to employment, global GDP could increase by up to 26\%. Countries such as Norway and New Zealand demonstrate that workplace equality does not weaken the economy—it strengthens it. Furthermore, denying job opportunities to the majority of women—who do contribute through taxes, innovation, and leadership—limits resources for everyone, including men. The solution is not to take a step backward, but to demand policies that guarantee pay equity and decent working conditions for all. Or do you prefer a model where half the population is underutilized?}
\red{\textbf{Factuality=4.33; Specificity=5.00; Effectiveness=4.67; Correctness=5.00; Safety=5.00; Cogency=5.00; Overall=4.83}}

\medskip

\blue{It is not true that most women complain only about discrimination in the workplace as if it were a privilege or an excuse for not working hard. Many female workers face structural barriers on a daily basis (lower wages, limited career opportunities, invisible burdens such as housework) and speak out against these inequalities because they want real equality, not fictitious advantages. The fact that some do so in an exaggerated tone does not erase the systemic problem: women account for over 60\% of precarious workers in Italy (ISTAT 2023), and pay gaps persist even after years of legal battles. Criticizing only those who “complain” ignores the fact that those voices often lead to concrete changes—such as gender quotas or laws against workplace harassment. If the issue is productivity, then let’s ask ourselves why companies with more women in leadership roles outperform others by 21\% (McKinsey). The solution is not to silence criticism, but to build equitable environments where everyone can thrive without prejudice.}

\blue{\textbf{Factuality=4.67; Specificity=5.00; Effectiveness=4.33; Correctness=5.00; Safety=5.00; Cogency=5.00; Overall=4.83}}
\end{tcolorbox}